\documentclass[final,5p,times,twocolumn,authoryear]{elsarticle}

\usepackage[utf8]{inputenc}
\usepackage[T1]{fontenc}
\usepackage{textcomp}
\usepackage{amsmath,amsfonts,amssymb}
\usepackage{booktabs}
\usepackage{nicefrac}
\usepackage{microtype}
\usepackage{tabularx}
\usepackage{array}
\usepackage{multirow}
\usepackage{makecell}
\usepackage{xcolor}
\usepackage{graphicx}
\usepackage{capt-of}
\usepackage{epstopdf}
\usepackage{xurl}
\usepackage{hyperref}

\newcommand{\best}{\textcolor{red}}
\newcommand{\second}{\textcolor{blue}}
\graphicspath{{./image/}}

\hypersetup{hidelinks,pdfauthor={Zekai Shi, Meng Zhang, Haokun Zhang, Bo Zhang},pdftitle={Efficient Continuous DEM Reconstruction under Limited Target-Resolution Supervision},pdfsubject={Preprint}}

\makeatletter
\let\ps@pprintTitle\ps@plain
\makeatother
\biboptions{round,semicolon}
\let\cite\citep

\begin{document}

\begin{frontmatter}

\title{Efficient Continuous DEM Reconstruction under Limited Target-Resolution Supervision}

\author[aff1]{Zekai Shi}
\author[aff1]{Meng Zhang\corref{cor1}}
\ead{zhangmeng01@mail.xjtu.edu.cn}
\author[aff1]{Haokun Zhang}
\author[aff2]{Bo Zhang}
\cortext[cor1]{Corresponding author.}

\affiliation[aff1]{organization={School of Human Settlements and Civil Engineering, Xi'an Jiaotong University},
            city={Xi'an}, postcode={710049}, country={China}}
\affiliation[aff2]{organization={School of Artificial Intelligence, Optics and Electronics (iOPEN), Northwestern Polytechnical University},
            city={Xi'an}, postcode={710072}, country={China}}

\begin{abstract}
High-resolution digital elevation models (DEMs) support Earth observation applications, but paired training references are often available only at coarser output resolutions. Reconstructing finer terrain grids therefore requires both effective transfer beyond the supervised scale and control of dense-query computation. To address this problem, SCOPE learns a continuous terrain representation from coarser-resolution pairs. It predicts a latent coefficient field on the low-resolution grid and reuses local Fourier residual functions through basis evaluation and geometry-guided ensemble fusion. This separates high-dimensional coefficient prediction from output-grid construction. Experiments on geographically distributed land--ocean samples assess supervised reconstruction, unseen-scale inference, cross-domain generalization, and theoretical computation. SCOPE leads the compared methods across six metrics in the main supervised-scale evaluation. At an unseen factor three times the training factor, land reconstruction reduces RMSE and MAE by approximately 12\% relative to bicubic interpolation, with errors close to target-scale fine-tuning. Ninefold output density increases counted multiply--accumulate operations by only about 2\%. Frozen-model validation on held-out external marine regions reduces RMSE relative to the DEM-specific implicit baseline EBCF-CDEM by approximately 19\% under self-downsampling and 2\% with cross-product inputs, while also yielding lower RMSE than LIIF-MS in both settings. These results demonstrate the value of reusable coefficient fields for accurate reconstruction beyond the supervised resolution with low incremental arithmetic cost.
\end{abstract}


\end{frontmatter}
\section{Introduction}
\label{sec:introduction}

\subsection{DEM Products and Limited Target-Resolution Supervision}

Digital elevation models (DEMs) and bathymetric grids are fundamental Earth observation products for terrain interpretation, hydrologic modeling, hazard assessment, coastal management, and marine resource exploration~\cite{millerDigitalTerrainModel1958,mooreDigitalTerrainModelling1991,smithGlobalSeaFloor1997}. The spatial resolution required by these applications is not always supported by the available observations. Satellite missions such as SRTM and TanDEM-X have improved terrestrial elevation coverage~\cite{farrShuttleRadarTopography2007,kriegerTanDEMXSatelliteFormation2007,wesselAccuracyAssessmentGlobal2018}, whereas marine observations remain more spatially incomplete. The GEBCO\_2024 release reports that 26.1\% of the global seabed had been mapped using modern survey techniques~\cite{gebco2024release}; extensive regions rely on bathymetric estimates inferred from satellite-altimetry-derived gravity anomalies~\cite{tozerGlobalBathymetryTopography2019,smithBathymetricPredictionDense1994}.

Airborne LiDAR, photogrammetry, interferometric SAR, and shipborne acoustic surveys can provide detailed terrain observations, but acquisition cost, sensor characteristics, and uneven coverage constrain their availability. Consequently, paired DEMs may support learning at a coarser output resolution even when reference data at the desired finer resolution are unavailable for training. The practical question is whether these available pairs can support reconstruction beyond interpolation at a finer target grid, without requiring new target-scale training labels. In this study, limited target-resolution supervision denotes learning from paired elevation references at a coarser output resolution than the desired reconstruction grid.

\subsection{Limitations of Fixed-Scale DEM Reconstruction}

Classical DEM enhancement uses interpolation, geostatistical estimation, multi-source fusion, and terrain examples to construct denser elevation grids~\cite{yangDeepLearningSingle2019,straubSuperresolutionTerrainMap2012,yueFusionMultiscaleDEMs2015}. Deep learning approaches have adapted convolutional, residual, adversarial, attention-based, and Transformer architectures to DEM reconstruction~\cite{dongLearningDeepConvolutional2014,xuDeepGradientPrior2019,demirayDSRGANSuperresolutionGenerative2021,zhouEnhancedDoublefilterDeep2021,zhangTerrainFeatureawareDeep2022}. DEM-specific models further exploit terrain decomposition, frequency separation, shaded relief, and hydrological or uncertainty-related information~\cite{wangSuperresolutionFrameworkBased2024,wenUnmixingFrequencyFeatures2025,huangSuperresolutionGuidedShaded2024,huangMultimodalSuperresolutionUsing2025}.

Many learned grid-to-grid approaches, including the fixed-scale configurations of EDSR, SwinIR, and HAT, optimize a mapping to a predefined output grid. Changing the reconstruction factor commonly requires a different output head or another training configuration with appropriate target-scale examples. Interpolation can produce arbitrary grid spacings without such training, but does not learn terrain-dependent corrections from paired observations. Thus, output-grid flexibility and learned elevation reconstruction need to be considered together when the desired resolution exceeds the available supervision.

\subsection{Continuous Terrain Representation}

Continuous representations offer a way to separate the output sampling grid from the learned terrain function. Coordinate-conditioned models evaluate elevations at requested spatial locations, while Fourier feature mappings and sinusoidal representations provide tools for representing spatial variation~\cite{tancikFourierFeaturesLet2020,sitzmannImplicitNeuralRepresentations2020}. Meta-SR, LIIF, LTE, and CiaoSR explore arbitrary-scale image reconstruction using scale-conditioned filters, local implicit functions, frequency-aware estimation, and neighborhood aggregation~\cite{huMetaSRMagnificationarbitraryNetwork2019,chenLearningContinuous2021,leeLocalTextureEstimator2022a,caoCiaoSRContinuousImplicit2023}. Continuous DEM and LIIF-style elevation models have likewise established arbitrary-coordinate terrain reconstruction~\cite{heSuperresolutionDigitalElevation2022,yaoContinuousDigitalElevation2024}.

These approaches share flexible coordinate querying but differ in the predicted quantities and their computational roles. LIIF predicts values from local features and query coordinates; LTE uses query-centered frequency-aware estimation; EBCF-CDEM predicts an elevation bias relative to a nearest-neighbor terrain base. These mechanisms provide relevant baselines for examining both reconstruction accuracy and the work repeated at each output coordinate. Continuous query capability alone does not establish the accuracy of a model beyond its training scales, and a larger number of output queries can also increase the repeated decoding cost.

\subsection{From Coordinate Querying to Reusable Terrain Functions}
\label{sec:research_gap}

To what extent can a continuous terrain representation learned under coarser-resolution supervision sustain reconstruction accuracy at finer, unseen scales without target-scale retraining or substantial growth in computational cost? The evaluation jointly examines reconstruction accuracy outside the supervised scale and computational growth as the output grid becomes denser.

To address this problem, SCOPE predicts a latent coefficient field on the low-resolution (LR) feature grid. The coefficients parameterize local Fourier residual functions that are reused across output coordinates. Local basis evaluation and a geometry-guided Local Attentive Ensemble (LAE) combine neighboring residual candidates over a bicubic base surface. This shifts high-dimensional coefficient prediction to the LR grid and leaves a lightweight evaluation task at the query stage. The central distinction is therefore the predicted object and the allocation of repeated computation, rather than the use of Fourier functions in isolation.

\begin{figure}[htbp]
\centering
\includegraphics[width=\columnwidth]{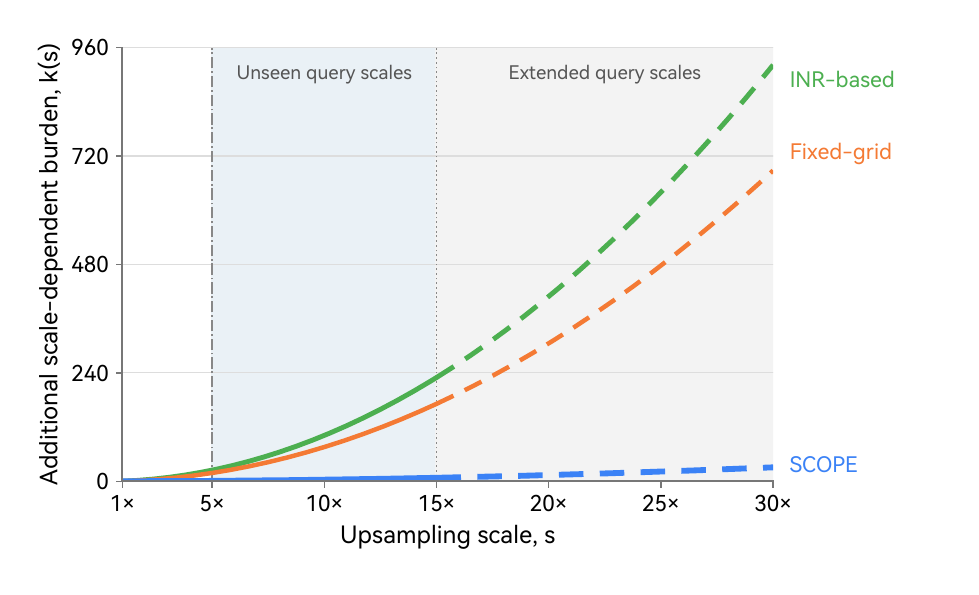}
\caption{Schematic architectural trends in scale-dependent computational burden from $1\times$ to $30\times$ for fixed-grid reconstruction, coordinate-wise INR decoding, and SCOPE. SCOPE concentrates coefficient prediction on the LR grid and uses lightweight local evaluation for denser output. Solid curves extend through $15\times$; dashed extensions illustrate conceptual trends to $30\times$. Markers identify the supervised $5\times$ scale and the maximum accuracy-evaluation factor of $15\times$. Quantitative GMACs and computation--latency trajectories are reported in Sec.~\ref{sec:discussion}.}
\label{fig:scale_compute_expansion}
\end{figure}

\begin{figure*}[htbp]
\centering
\includegraphics[width=0.95\textwidth]{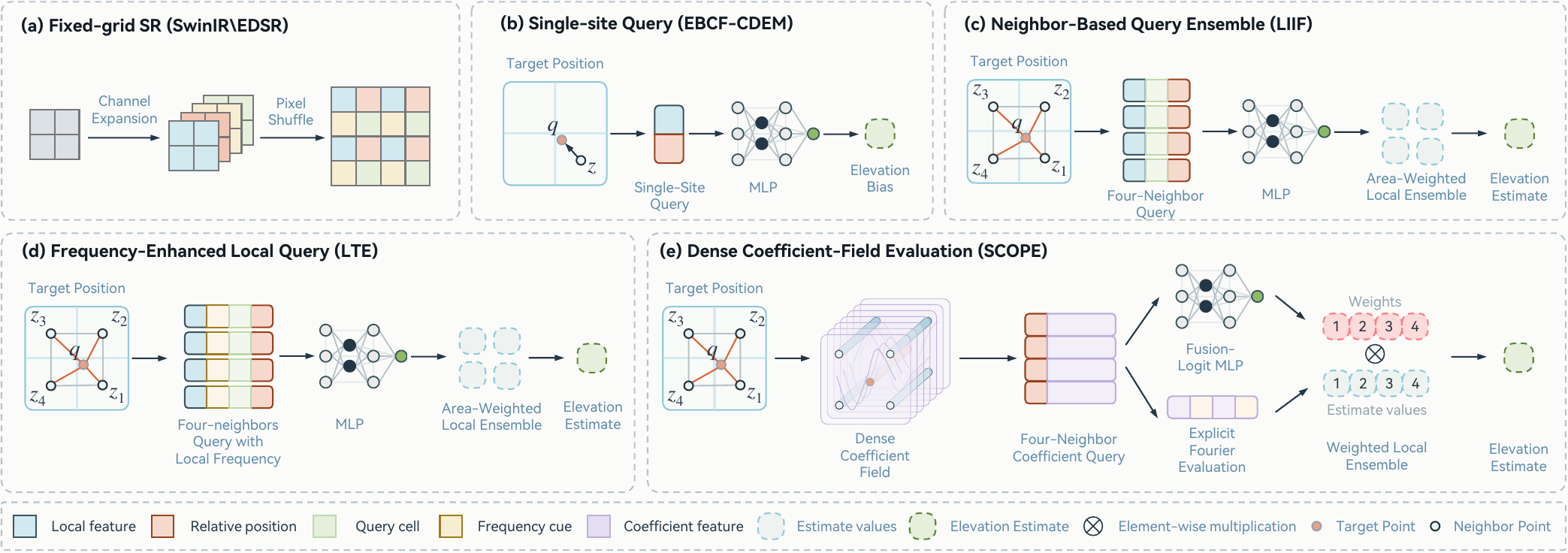}
\caption{Conceptual distinction among representative continuous reconstruction methods and SCOPE. The comparison highlights the predicted object, supervision setting, and decoding mechanism: query-wise implicit value prediction, query-centered frequency-aware reconstruction, DEM-specific elevation-bias prediction, and reusable latent coefficient-field evaluation.}
\label{fig:conceptual_comparison}
\end{figure*}

Figure~\ref{fig:scale_compute_expansion} illustrates the output-scale motivation while retaining fixed-grid and coordinate-wise INR reconstruction as separate reference categories. Figure~\ref{fig:conceptual_comparison} summarizes the corresponding representation and decoding mechanisms.

The main contributions are summarized as follows:
\begin{itemize}
\item A reconstruction study under limited target-resolution supervision examines whether a model learned from coarser LR--HR pairs can reconstruct a finer target grid without target-scale retraining, distinguishing output-grid flexibility from demonstrated scale extrapolation.
\item A reusable coefficient-field formulation concentrates high-dimensional prediction on the LR grid and reconstructs elevation residuals through local basis evaluation and geometry-guided fusion, reducing the arithmetic work repeated across dense output coordinates.
\item An evaluation combines supervised-scale accuracy, unseen-scale reconstruction, and theoretical operation counts with external cross-domain tests. The external tests distinguish cross-region generalization under synthetic degradation from transfer to matched inputs drawn from a different DEM product.
\end{itemize}

The remainder of this paper is organized as follows. Section~\ref{sec:data} describes the data sources, study design, sample construction, and evaluation protocol. Section~\ref{sec:method} presents the SCOPE network. Section~\ref{sec:experiment} reports the experimental setup, baseline comparison, ablation studies, and qualitative analyses. Section~\ref{sec:discussion} discusses implications and limitations. Section~\ref{sec:conclusion} concludes the paper.

\section{Data and Study Design}
\label{sec:data}

\subsection{Data Sources}
\label{sec:data_sources}

Two complementary data collections are used: a main land--ocean collection for supervised reconstruction and scale-transfer experiments, and a held-out marine collection for frozen-model external evaluation. Table~\ref{tab:dem_products} summarizes their source products and experimental roles. Figure~\ref{fig:study_area} maps source-product coverage and regional extents, while Secs.~\ref{sec:preprocessing} and~\ref{sec:external_data} describe sample construction and the evaluated subsets.

The elevation products integrated in this study include global terrestrial DEMs, global background bathymetry, and regional high-resolution coastal or marine datasets. Their main spatial characteristics are summarized in Table~\ref{tab:dem_products}. For products originally distributed on angular grids, spatial resolutions are reported as approximate meter-scale ground spacing to facilitate comparison with the 30~m reference products. Specifically, the 1 arc-second NOAA CRM grid corresponds to approximately 30~m, the 15 arc-second GEBCO\_2024 grid corresponds to approximately 450~m, and the 1/16 arc-minute EMODnet DTM grid corresponds to approximately 100~m near the equator. These values represent nominal ground spacing converted from the native angular grids; the actual east--west ground spacing varies with latitude.

TanDEM-X 30~m EDEM is used as the primary high-resolution terrestrial reference because it provides a globally consistent radar-derived representation of land topography with fine spatial resolution and homogeneous acquisition characteristics \cite{dlr_tandemx_30m_edem,rizzoliGenerationPerformanceAssessment2017}. GEBCO\_2024 serves as the global ocean and background bathymetric product, providing broad terrain coverage in marine regions where direct high-resolution acoustic observations are sparse \cite{gebcocompilationgroupGEBCO2024Grid2024}.

\begin{table*}[htbp]
\centering
\caption{DEM and bathymetric source products and their experimental roles. Parenthetical region identifiers refer to Fig.~\ref{fig:study_area}. The first six products support the main study; GEBCO\_2023 and the four regional products below it support frozen-model external evaluation. All external reference rasters are prepared on a nominal 90~m evaluation grid. The external subset contains 12 reference rasters and 3,825 valid patches; Great Barrier Reef blocks B--D are used, with block A excluded for spatial overlap with the main-study region.}
\label{tab:dem_products}
\scriptsize
\setlength{\tabcolsep}{2.5pt}
\renewcommand{\arraystretch}{1.15}
\begin{tabular*}{\textwidth}{@{\extracolsep{\fill}}llcl@{}}
\toprule
\textbf{Product} & \textbf{Coverage} & \textbf{Grid spacing} & \textbf{Role in this study} \\
\midrule
TanDEM-X EDEM & Global land & 30~m & Land HR reference \\
GEBCO\_2024 & Global ocean and land & $\sim$450~m & Ocean background \\
NOAA CRMs (A--C, E) & U.S. coastal zones & $\sim$30~m & Land--sea transition reference \\
Torres Strait 2023 (F) & \makecell[l]{139--146$^\circ$E, 8--13$^\circ$S} & 30~m & Reef and island reference \\
Bass Strait 2022 (G) & \makecell[l]{143--149$^\circ$E, 38--41$^\circ$S} & 30~m & Shelf bathymetry reference \\
EMODnet DTM 2024 (D) & Caribbean tile & $\sim$100~m & Fixed-$5\times$ evaluation only \\
\midrule
GEBCO\_2023 & Global ocean and land & $\sim$450~m & Matched cross-product LR input \\
CHS NONNA-100 (H$_1$, H$_2$) & Canadian waters & $0.001^\circ$ (source) & External bathymetric reference \\
MH370 bathymetry (I) & Southern Indian Ocean survey areas & 90~m (evaluation) & External bathymetric reference \\
AusBathyTopo Northern Australia (J) & \makecell[l]{121--133$^\circ$E, 18--8$^\circ$S} & 30~m (source) & External bathymetric reference \\
AusBathyTopo Great Barrier Reef (K) & Northeastern Australian waters & 30~m (source) & External reference, blocks B--D \\
\bottomrule
\end{tabular*}
\end{table*}

Four regional bathymetric products are further incorporated to improve the representation of high-resolution coastal and marine terrain. The refined 1 arcsec NOAA Coastal Relief Models (CRMs) from Vols.~1--5, 7, 9 and 10 provide $\sim$30~m topographic--bathymetric surfaces for U.S. coastal zones and characterize land--ocean transition terrain \cite{noaa_crm_2023}. The AusBathyTopo Torres Strait 30~m 2023 and Bass Strait 30~m 2022 datasets provide representative Australian reef--island and shallow-shelf bathymetry, respectively, with complex local gradients and smoother shelf backgrounds \cite{geoscience_australia_torres_strait_2023,geoscience_australia_bass_strait_2022}.

The EMODnet Digital Bathymetry DTM 2024 Caribbean tile further expands the regional marine sample set at approximately 100~m resolution by integrating survey, acoustic, satellite-derived, and global background bathymetric sources \cite{emodnet_bathymetry_2024}. After resampling, this dataset is used only in the fixed $5\times$ evaluation subset. Together, these regional products complement TanDEM-X and GEBCO\_2024 by adding coastal, reef, shallow-shelf, and open-marine terrain conditions.

\subsection{Study Area and Sample Construction}
\label{sec:preprocessing}

\begin{figure*}[htbp]
\centering
\includegraphics[width=0.95\textwidth]{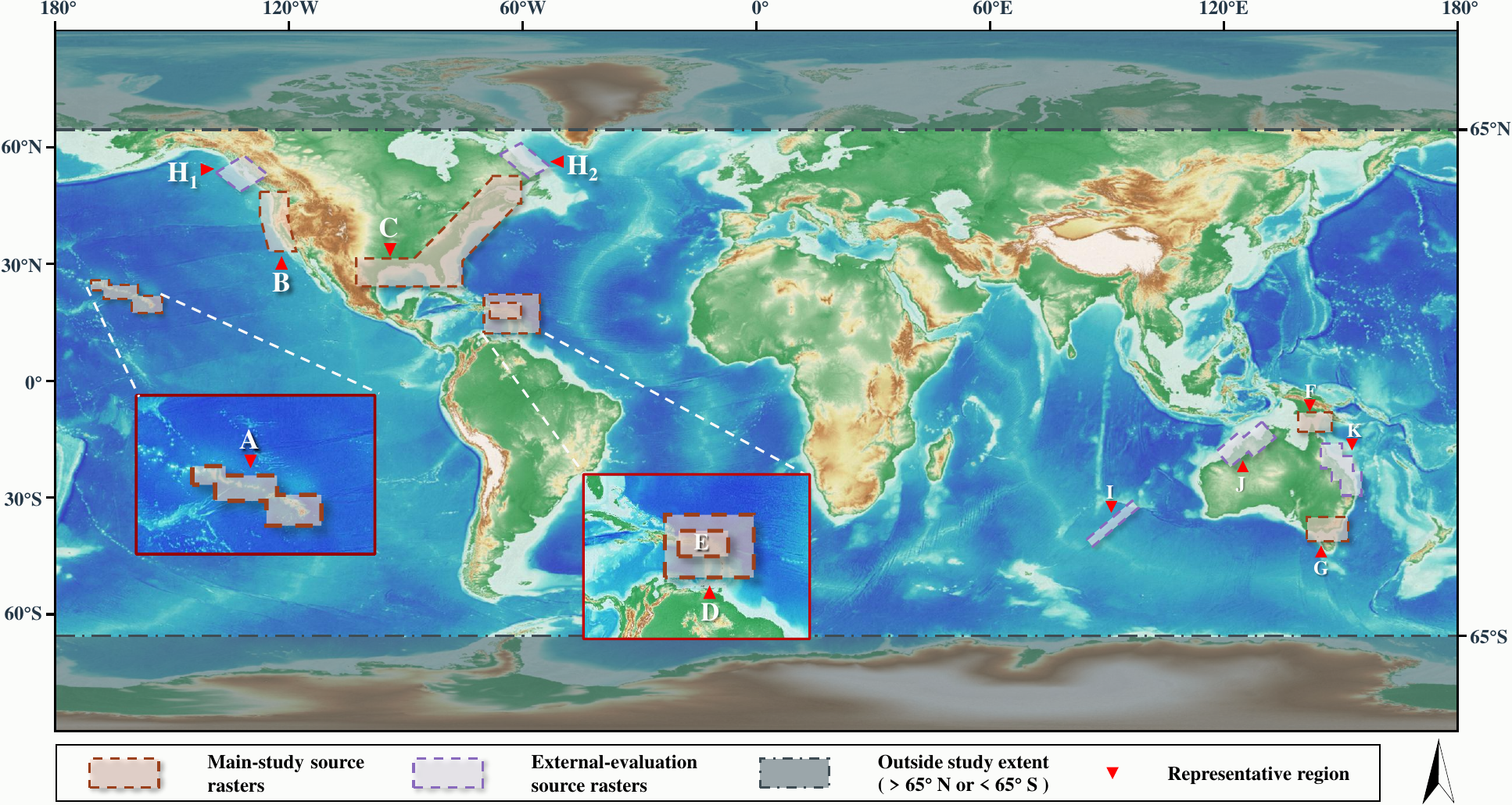}
\caption{Geographic distribution and coverage context of the source products. Brown and purple outlines distinguish main-study and external-evaluation product extents, respectively. A, B, C, and E identify NOAA CRMs Vols.~10, 7, 1--5, and 9; D identifies the EMODnet DTM 2024 Caribbean tile; F and G identify Torres Strait 2023 and Bass Strait 2022. H$_1$/H$_2$, I, J, and K identify CHS NONNA-100, MH370 bathymetry, Northern Australia, and Great Barrier Reef sources, respectively. The evaluated subsets are defined in Table~\ref{tab:dem_products} and Sec.~\ref{sec:external_data}. Red triangles locate representative regions, insets enlarge selected areas, and gray shading marks latitudes outside the $65^\circ$N--$65^\circ$S study limits.}
\label{fig:study_area}
\end{figure*}

The study encompasses geographically distributed terrain samples between 65\textdegree{}N and 65\textdegree{}S. Figure~\ref{fig:study_area} distinguishes source-product coverage associated with the main study from that associated with external marine evaluation. The selected extent covers complex land, coastal, and shallow-marine terrain, providing the spatial basis for constructing multi-source land--ocean terrain samples.

Both terrestrial and marine DEM products are included during sample construction. The land DEM samples provide relatively abundant and structurally clear high-resolution terrain references, including elevation variation, slope transitions, ridge--valley morphology, and local relief patterns. Marine and coastal samples complement the land data with bathymetric surfaces and land--ocean transition structures, where high-resolution observations are usually sparser and more heterogeneous.

All DEM and bathymetric products are processed under a unified GDAL-based geospatial workflow before sample construction. The original datasets are transformed to the WGS84 geographic coordinate system, converted to meter-based elevation or depth units when necessary, and standardized into GeoTIFF raster format. Nodata regions, abnormal elevation values, and invalid bathymetric cells are removed or excluded during preprocessing.

For paired or cross-product comparisons, LR and HR rasters are further co-registered and aligned on the raster grid before patch extraction. When products have different native resolutions or grid definitions, resampling is performed within the same GDAL-based workflow to ensure that paired samples correspond to the same geographic footprint. This alignment step reduces spatial mismatch among heterogeneous DEM products and provides spatially consistent patches for subsequent supervised reconstruction and evaluation.
The evaluation patches were generated through a predefined automated sampling and terrain-classification procedure without manual location selection.

This study employs two types of LR--HR sample construction. The first is controlled paired construction, where the LR input is generated from the HR reference through a predefined degradation operator:
\begin{equation}
I_{\mathrm{LR}}=\mathcal{D}_{s}(I_{\mathrm{HR}}).
\label{eq:controlled_pair}
\end{equation}
Here, $\mathcal{D}_{s}(\cdot)$ denotes the scale-dependent degradation process and $s$ is the reconstruction scale. This setting provides reproducible LR--HR pairs under controlled resolution relationships.

The second is cross-product matched construction, where the LR and HR samples are obtained from different elevation products:
\begin{equation}
(I_{\mathrm{LR}}^{m},I_{\mathrm{HR}}^{m})
=
\mathcal{M}(I_{\mathrm{LR}}^{raw},I_{\mathrm{HR}}^{raw}).
\label{eq:matched_pair}
\end{equation}
Here, $\mathcal{M}(\cdot)$ represents spatial co-registration, resampling, valid-region intersection, and patch-level quality control. This setting represents paired samples in which the LR input and HR reference are derived from different elevation products.

The resulting paired and cross-product samples provide the data basis for the experimental evaluation described in Sec.~\ref{sec:experiment}.

\subsection{External Marine Data for Cross-Domain Evaluation}
\label{sec:external_data}

External marine evaluation uses four regional source groups, identified as H$_1$/H$_2$--K in Fig.~\ref{fig:study_area}. CHS NONNA-100 provides Canadian bathymetry distributed in geographic tiles; south of $68^\circ$N, the nominal source grid spacing is $0.001^\circ$~\cite{chs_nonna_bathymetry}. The MH370 data provide bathymetric survey areas and track coverage in the southern Indian Ocean~\cite{ga_mh370_release}. The Northern Australia 30~m product covers the northwestern--northern shelf within 121--133$^\circ$E and 18--8$^\circ$S~\cite{ga_northern_australia_2018}. The Great Barrier Reef 30~m product comprises overlapping regional blocks; B--D enter this evaluation, with A excluded because of spatial overlap with the main-study region~\cite{ga_great_barrier_reef_depth_model}.

The selected subset contains 12 reference rasters, all prepared on a nominal 90~m evaluation grid from the source products listed in Table~\ref{tab:dem_products}. These rasters are held out from model training and checkpoint selection. They yield 3,825 valid patches: 18 from the Canadian group, 27 from MH370, 1,116 from Northern Australia, and 2,664 from the Great Barrier Reef. Aggregate metrics use equal weighting of valid patches.

Two LR constructions use the same reference patches and nominal $450\rightarrow90$~m reconstruction factor. In the self-downsampled setting, bicubic downsampling of the 90~m reference generates the 450~m LR input, testing cross-region transfer under a controlled synthetic degradation. In the matched cross-product setting, the LR input is drawn from GEBCO\_2023 and aligned with the regional reference~\cite{thenipponfoundation-gebcoGEBCO_2023Grid2023}. Each setting is evaluated and reported separately.

\section{Methodology}
\label{sec:method}

This section describes the proposed SCOPE network for continuous DEM reconstruction. As illustrated in Fig.~\ref{fig:scope_framework}, the SCOPE network comprises a terrain encoder, the Neighborhood-Attentive Field (NAF) decoder, the Local Attentive Ensemble (LAE) module, and a bicubic base surface. The LR DEM is first encoded into an LR-aligned terrain feature field and transformed into a latent coefficient field. For a continuous query coordinate $\mathbf{q}$, NAF samples neighboring coefficient vectors and evaluates local residual candidates $r_i(\mathbf{q})$ through local basis evaluation. LAE estimates compatibility weights among these candidates and fuses them into $r(\mathbf{q})$, which is added to the bicubic base surface $B(\mathbf{q})$ to produce the reconstructed elevation $\widehat{z}(\mathbf{q})$.

The following subsections introduce the overall SCOPE network, the NAF decoder and its latent coefficient field, the LAE module, the activation function, and the training objective together with the evaluation protocol.

\subsection{SCOPE Network}
\label{sec:scope_framework}

\begin{figure*}[htbp!]
\centerline{\includegraphics[width=0.95\textwidth]{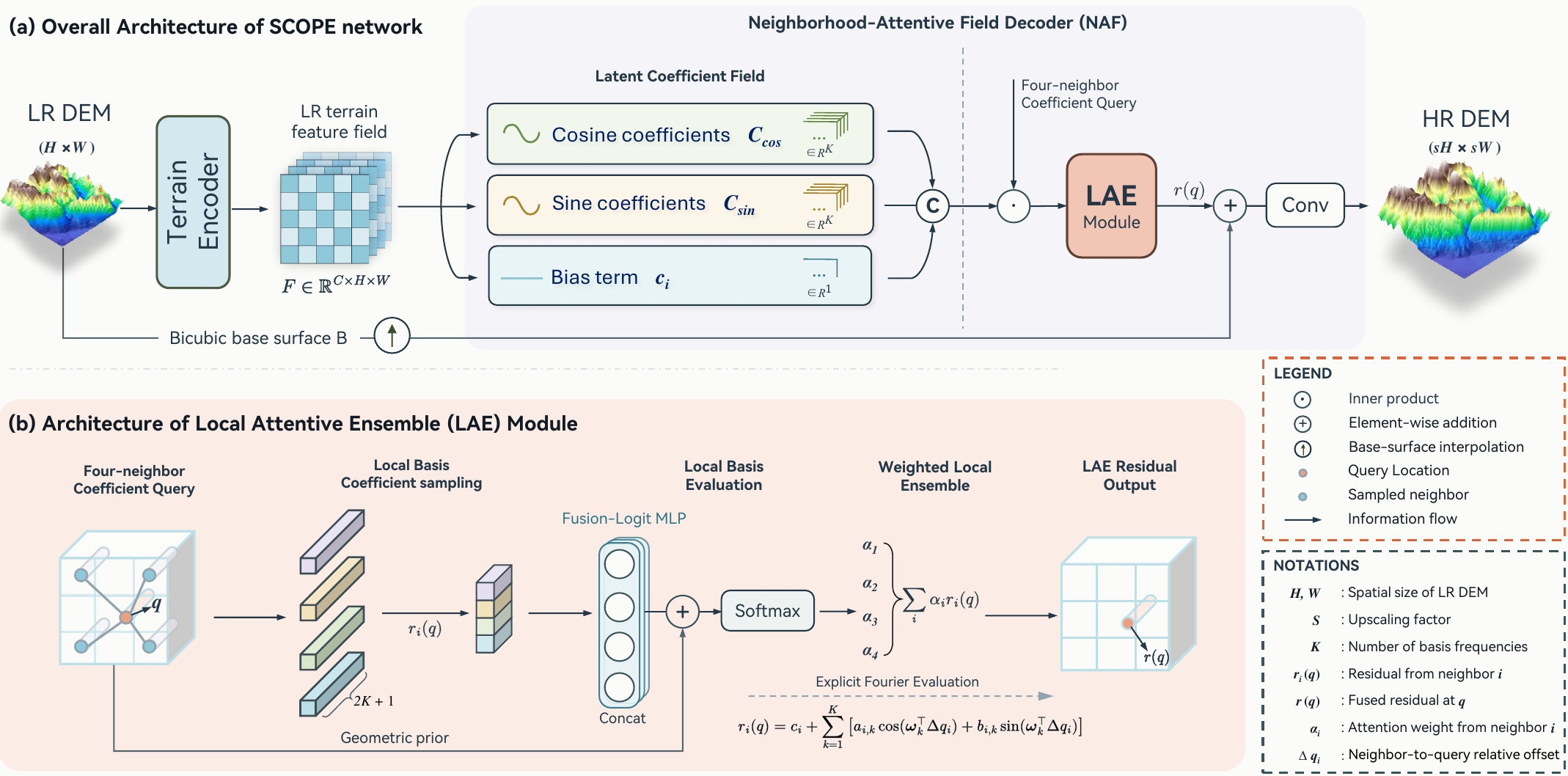}}
\caption{Architecture of SCOPE network.
(a) Overall framework. The LR DEM is encoded into an LR-aligned terrain
feature field and transformed by NAF into a latent coefficient field.
For a continuous query coordinate $\mathbf{q}$, coefficients sampled
at neighboring LR locations $\mathbf{x}_{i}$ are evaluated using the
relative offsets $\Delta\mathbf{q}_{i}$ to produce local residual
candidates $r_i(\mathbf{q})$. These candidates are fused by LAE into
$r(\mathbf{q})$ and added to the bicubic base surface $B(\mathbf{q})$.
(b) LAE module. Four neighboring residual candidates are fused using
geometry-guided compatibility weights.}
\label{fig:scope_framework}
\end{figure*}

As illustrated in Fig.~\ref{fig:scope_framework}, SCOPE reconstructs a continuous HR elevation surface through a base--residual formulation. The bicubic branch provides a conservative base surface $B(\mathbf{q})$ that preserves the large-scale elevation trend of the LR DEM, while the Neighborhood-Attentive Field (NAF) decoder predicts a latent coefficient field from the LR terrain context. Neighboring coefficient vectors define local residual candidates $r_i(\mathbf{q})$ at the query coordinate $\mathbf{q}$, and the Local Attentive Ensemble (LAE) fuses these candidates into $r(\mathbf{q})$ before they are added back to the base surface. Following their initial definition, the abbreviations NAF and LAE denote these components throughout this section.

Let $\Omega\subset\mathbb{R}^{2}$ denote the continuous spatial domain of the target DEM and let $\Lambda_{\mathrm{LR}}\subset\Omega$ denote the LR sampling lattice. Given an LR DEM patch $I_{\mathrm{LR}}$ and a continuous query coordinate $\mathbf{q}\in\Omega$, SCOPE defines the reconstructed elevation as
\begin{equation}
\widehat{z}(\mathbf{q})
=
B(\mathbf{q})
+
r(\mathbf{q})
\label{eq:scope_residual}
\end{equation}
where $B(\mathbf{q})$ is the bicubic base surface evaluated at $\mathbf{q}$ and $r(\mathbf{q})$ is the LAE-fused residual. This formulation separates low-frequency terrain continuity from fine-scale residual reconstruction, so the query coordinate $\mathbf{q}$ is used to evaluate local residual functions encoded in the coefficient field rather than to drive direct coordinate-to-elevation regression.

The input to SCOPE combines the normalized LR DEM with an amplitude-normalization condition channel. The terrain encoder maps this input to an LR-aligned feature field:
\begin{equation}
\mathbf{F}
=
\mathcal{E}_{\theta}(I_{\mathrm{LR}},C_n),
\label{eq:terrain_feature}
\end{equation}
where $C_n$ denotes the amplitude-normalization condition field and $\mathcal{E}_{\theta}$ is the terrain encoding operator. Under the global normalization used here, $C_n$ is constant; the output sampling density is specified through the query grid. The feature field $\mathbf{F}$ remains aligned with the LR lattice $\Lambda_{\mathrm{LR}}$, because the encoder functions as a context extractor rather than a grid upsampling module. This alignment allows each LR lattice point to provide local terrain context for coefficient prediction and continuous residual evaluation.

NAF maps the LR-aligned feature field to a latent coefficient field instantiated by local basis coefficients:
\begin{equation}
\boldsymbol{\Theta}
=
\mathcal{G}_{\phi}(\mathbf{F})
=
\left(
\mathcal{C},
\{\mathcal{A}_{k},\mathcal{B}_{k}\}_{k=1}^{K}
\right),
\label{eq:coefficient_field}
\end{equation}
where $\mathcal{G}_{\phi}$ denotes the coefficient prediction operator, $\mathcal{C}$ is the bias coefficient field, and $\mathcal{A}_{k}$ and $\mathcal{B}_{k}$ are the cosine and sine coefficient fields associated with the $k$-th basis component. These channels form an LR-lattice-aligned latent coefficient field whose sampled vectors parameterize local residual functions rather than directly storing discrete elevation residual values.

For the $i$th neighboring LR location $\mathbf{x}_{i}\in\Lambda_{\mathrm{LR}}$, the local coefficient vector is obtained by sampling the coefficient fields at $\mathbf{x}_{i}$:
\begin{equation}
\begin{aligned}
\boldsymbol{\theta}_{i}
&=
\left(c_{i},\{a_{i,k},b_{i,k}\}_{k=1}^{K}\right),\\
&=
\left(
\mathcal{C}(\mathbf{x}_{i}),
\{\mathcal{A}_{k}(\mathbf{x}_{i}),
\mathcal{B}_{k}(\mathbf{x}_{i})\}_{k=1}^{K}
\right).
\end{aligned}
\label{eq:local_coefficient_vector}
\end{equation}
where $c_{i}=\mathcal{C}(\mathbf{x}_{i})$, $a_{i,k}=\mathcal{A}_{k}(\mathbf{x}_{i})$, and $b_{i,k}=\mathcal{B}_{k}(\mathbf{x}_{i})$ denote the bias, cosine, and sine coefficients sampled at the $i$th neighboring LR location $\mathbf{x}_{i}$, respectively. For an arbitrary continuous query coordinate $\mathbf{q}$, the relative displacement with respect to the $i$th neighboring LR location $\mathbf{x}_{i}$ is defined as $\Delta\mathbf{q}_{i}=\mathbf{q}-\mathbf{x}_{i}$.
Rather than directly regressing the elevation at $\mathbf{q}$, NAF evaluates a local residual candidate associated with $\mathbf{x}_{i}$ from the sampled coefficient vector as
\begin{equation}
\begin{aligned}
r_{i}(\mathbf{q})
&=c_{i}+\sum_{k=1}^{K}\Bigl[
a_{i,k}\cos\left(
\boldsymbol{\omega}_{k}^{\mathrm{T}}\Delta\mathbf{q}_{i}
\right)\\
&\hspace{3.9em}+
b_{i,k}\sin\left(
\boldsymbol{\omega}_{k}^{\mathrm{T}}\Delta\mathbf{q}_{i}
\right)\Bigr],
\end{aligned}
\label{eq:fourier_residual}
\end{equation}
where $\boldsymbol{\omega}_{k}=2\pi\mathbf{f}_{k}$ is the $k$-th angular wave vector. This equation defines a local residual surface around each neighboring LR location. Therefore, changing the density of query coordinates produces DEM outputs at different reconstruction scales by repeatedly evaluating the same latent coefficient field, without changing the encoded feature field or introducing a scale-specific output layer.

The local residual candidates from neighboring LR locations are combined by LAE. For a continuous query coordinate $\mathbf{q}$, let $\mathcal{N}(\mathbf{q})=\{\mathbf{x}_{i}\}_{i=1}^{4}$ denote the four surrounding LR locations. Each neighboring LR location $\mathbf{x}_{i}\in\mathcal{N}(\mathbf{q})$ provides a local residual candidate $r_{i}(\mathbf{q})$, a relative displacement $\Delta\mathbf{q}_{i}$, and a bilinear geometric weight $w_i^{\mathrm{bilinear}}$. LAE computes a geometry-guided compatibility logit and normalized ensemble weight as
{\setlength{\jot}{6pt}
\begin{align}
\ell_i(\mathbf{q})
&=
g_{\psi}
\left(
r_{i}(\mathbf{q}),
\Delta\mathbf{q}_{i},
w_i^{\mathrm{bilinear}}
\right)\notag\\[-2pt]
&\quad+
\log\left(w_i^{\mathrm{bilinear}}+\epsilon\right)
\label{eq:lae_logit}\\
\alpha_i(\mathbf{q})
&=
\frac{\exp(\ell_i(\mathbf{q}))}
{\sum_{j=1}^{4}\exp(\ell_j(\mathbf{q}))},
\label{eq:lae_weight}\\
r(\mathbf{q})
&=
\sum_{i=1}^{4}
\alpha_i(\mathbf{q})r_i(\mathbf{q}),
\qquad
\sum_{i=1}^{4}\alpha_i(\mathbf{q})=1.
\label{eq:lae_fusion}
\end{align}}
Here, $g_{\psi}$ is a lightweight MLP, $\ell_i(\mathbf{q})$ is the compatibility logit, $\alpha_i(\mathbf{q})$ is the LAE weight, and $\epsilon$ is a small constant for numerical stability. The learned score estimates the relative compatibility of each local residual candidate $r_i(\mathbf{q})$ with the query location, while the logarithmic bilinear term injects the geometric interpolation prior into the logit space. As a result, LAE retains the spatial preference of bilinear interpolation while allowing terrain-dependent adjustment around local structures.

Finally, the LAE-fused residual $r(\mathbf{q})$ in Eq.~\eqref{eq:lae_fusion} is substituted into the base--residual reconstruction in Eq.~\eqref{eq:scope_residual}. These operations define the continuous base--residual representation. In the full implementation, the residual is scaled by $\eta$ before addition, giving $z_0(\mathbf{q})=B(\mathbf{q})+\eta r(\mathbf{q})$. For a complete output grid, a lightweight convolutional refinement $\mathcal{R}$ produces $\widehat{I}=I_0+\eta\mathcal{R}(I_0)$, where $I_0$ samples $z_0$ on that grid. The query coordinate $\mathbf{q}$ is not directly mapped to an elevation value. Instead, coefficient vectors sampled from neighboring LR locations are evaluated using the relative displacements $\Delta\mathbf{q}_{i}$, producing local residual candidates $r_{i}(\mathbf{q})$ that are subsequently fused by LAE.

\subsection{Sign-Adaptive Smooth Unit}
\label{sec:sasu}

The SCOPE network is required to model both positive and negative terrain residual responses. Positive residual responses may correspond to locally sharpened ridges, raised terrain details, or seabed features that are smoothed in the coarse DEM, whereas negative responses may correspond to incised channels, valleys, depressions, or local bathymetric troughs. To improve nonlinear residual representation, a parameter-free sign-adaptive activation function, termed the Sign-Adaptive Smooth Unit (SASU), is adopted in the elevation residual decoder.

For an input activation value $x$, SASU combines the Gaussian error linear unit (GELU) and the sigmoid-weighted linear unit (SiLU) in a sign-adaptive piecewise manner:
{\setlength{\jot}{6pt}
\begin{align}
\mathrm{GELU}(x) &= x\Phi(x)=\frac{x}{2}\left[1+\operatorname{erf}\left(\frac{x}{\sqrt{2}}\right)\right], \nonumber\\
\mathrm{SiLU}(x) &= x\sigma(x)=\frac{x}{1+\exp(-x)}, \nonumber\\
\mathrm{SASU}(x) &=
\begin{cases}
\mathrm{GELU}(x), & x \geq 0,\\
\mathrm{SiLU}(x), & x < 0.
\end{cases}
\label{eq:sasu}
\end{align}
}
where $\Phi(\cdot)$ denotes the cumulative distribution function of the standard normal distribution, $\sigma(\cdot)$ denotes the sigmoid function, and $\operatorname{erf}(\cdot)$ denotes the error function. The two branches share the same function value and first-order derivative at the origin, i.e., $\mathrm{SASU}(0)=0$ and $\mathrm{SASU}'(0)=1/2$, ensuring a smooth transition during optimization.

The positive GELU branch provides smooth activation while preserving salient positive terrain responses, whereas the negative SiLU branch maintains non-zero responses and gradients for suppressed features. Compared with ReLU, negative features are not hard-truncated; compared with applying GELU or SiLU alone, positive and negative residual responses are modulated by different nonlinear behaviors. SASU serves as the default activation function in SCOPE, especially in the elevation residual decoder.

\subsection{Training Objective}
\label{sec:training_objective}

The SCOPE network is trained with paired LR and HR DEM samples. The objective combines reconstruction, gradient, and circular direction losses:
{\setlength{\jot}{6pt}
\begin{align}
\mathcal{L}_{\mathrm{rec}}&=
\left\|I_{\mathrm{SR}}-I_{\mathrm{HR}}\right\|_{1},\\
\mathcal{L}_{\mathrm{grad}}&=
\left\|\nabla I_{\mathrm{SR}}-\nabla I_{\mathrm{HR}}\right\|_{1},\\
\mathcal{L}_{\mathrm{deg}}&=
\frac{\sum_{p}m_p\left[1-\operatorname{clip}(c_p,-1,1)\right]}{\sum_{p}m_p},\\
\mathcal{L}_{\mathrm{total}}&=
\lambda_{1}\mathcal{L}_{\mathrm{rec}}+
\lambda_{2}\mathcal{L}_{\mathrm{grad}}+
\lambda_{3}\mathcal{L}_{\mathrm{deg}} .
\label{eq:total_loss}
\end{align}
}
where $\nabla(\cdot)$ denotes the spatial gradient operator and $\lambda_{1}$, $\lambda_{2}$, and $\lambda_{3}$ are loss weights. For each pixel $p$, $c_p$ is the numerically stabilized cosine similarity between the SR and HR Sobel-gradient vectors. The mask $m_p$ selects HR gradient magnitudes above a per-sample relative threshold; the direction loss is zero when no valid pixels remain. This circular loss constrains local orientation without imposing an artificial discontinuity at the angular wraparound. The reconstruction and gradient terms preserve elevation consistency and local relief variation.

The complete experimental protocol, including same-scale reconstruction, ablation analysis, unseen-scale inference, scale-transfer evaluation, and terrain-stratified assessment, is described in Sec.~\ref{sec:experiment}.

\section{Experimental Setup and Results}
\label{sec:experiment}

This section evaluates reconstruction under limited target-resolution supervision through component ablation, fixed-scale accuracy, terrain-structure assessment, cross-domain evaluation, and unseen-scale reconstruction. Cross-domain evaluation separates the existing paired-product experiment from frozen-model tests on external marine regions.

\subsection{Experimental Design}
\label{sec:experimental_design}

The experiments evaluate SCOPE in terms of supervised-scale reconstruction, reconstruction beyond the supervised scale, terrain-structure preservation, and external transfer. The data construction protocol and study regions follow Sec.~\ref{sec:data}. Training and validation regions are spatially separated before patch extraction, with no overlapping or adjacent patches across the two subsets. Gradient updates use the training subset; the validation subset supports training monitoring, checkpoint selection, and the main quantitative comparisons. The external marine data in Sec.~\ref{sec:external_data} are reserved for frozen-model evaluation, with checkpoints fixed before external testing.

Unless otherwise specified, experiments are conducted under fixed $5\times$ supervised reconstruction. The physical-resolution protocol uses a nominal 450~m input grid and 90~m supervision. The fixed-$5\times$ model is subsequently evaluated at a nominal 30~m output grid ($15\times$) over the same spatial footprint. SCOPE-5F uses 90~m training labels; 30~m target-resolution supervision is supplied to the explicitly identified SCOPE-15D and SCOPE-15FT controls. Native source-product resolution and training-label grid spacing are specified separately. LR inputs for training and validation are generated from HR DEMs by bicubic downsampling with antialiasing, and the validation scale remains fixed at $5\times$. Elevation values are globally normalized after excluding the most extreme 0.1\% of values. Land and marine samples are jointly used during training at a target ratio of 3:1, and all compared methods follow the same fixed-$5\times$ preprocessing and evaluation protocol. The $15\times$ evaluation uses the land portion of the fixed $5\times$ evaluation data.

SCOPE is implemented as a single-encoder continuous DEM reconstruction model. The terrain encoder follows a shifted-window Transformer design and is trained end-to-end within SCOPE to extract 192-dimensional terrain features. The Neighborhood-Attentive Field (NAF) decoder predicts latent coefficient fields instantiated with 96 basis frequencies. Four neighboring local residual candidates are fused by the Local Attentive Ensemble (LAE), added to the bicubic base surface with a residual scaling factor of 0.1, and passed through the post-refinement module in the full configuration. SASU is used as the default activation function.

Model optimization follows an iteration-based protocol over approximately 230,000 LR--HR DEM patch pairs. Adam is used with a batch size of 6, an initial learning rate of $1\times10^{-4}$, a minimum learning rate of $1\times10^{-6}$, and a weight decay of $1\times10^{-6}$. A cosine learning-rate schedule and gradient clipping with a maximum norm of 2.0 are adopted. Validation is performed every 2000 iterations, and the checkpoint with the best validation composite score is selected for the reported quantitative comparisons.

The training objective combines elevation reconstruction, gradient, and aspect losses with weights of 1.0, 0.05, and 0.01, respectively. Edge loss is disabled, and the aspect loss is computed only in valid sloped regions to avoid unstable directional supervision over nearly flat terrain.

The compared methods include interpolation baselines, fixed-scale deep super-resolution models, and implicit neural reconstruction models. Learning-based baselines follow the same LR--HR data protocol whenever applicable. RMSE and MAE measure physical elevation errors in meters, while PSNR is reported in decibels. The table headings Slope, Aspect, and Corr. consistently denote the logged RMSE-Slope, RMSE-Aspect, and SlopeCorr metrics, respectively; Corr. measures slope consistency between reconstructed and reference terrain structures. Slope-related metrics are calculated with unit pixel spacing and quantify relative grid-space structure. Aspect errors use circular angular differences with a slope-based validity mask under the same convention. Lower RMSE, MAE, Slope, and Aspect and higher PSNR and Corr. indicate better performance.

PSNR is computed on globally normalized elevation arrays clipped to $[0,1]$, using a unit data range.

For the external evaluation, the existing SCOPE-5F, EDSR, LIIF-MS, and EBCF-CDEM checkpoints are frozen. Every method is evaluated on the same 3,825 valid reference patches in each input setting, with bicubic interpolation retained as the non-learned baseline. RMSE and MAE are calculated per patch in the physical elevation domain in meters, and their arithmetic means are reported over all valid patches. Relative error reduction is $100(E_{\mathrm{bicubic}}-E_{\mathrm{method}})/E_{\mathrm{bicubic}}$, so negative values indicate higher error. The final scale-transfer experiment compares unseen-scale inference, direct $15\times$ training, and $5\times$ training followed by $15\times$ fine-tuning to examine whether fixed $5\times$ supervision can support reconstruction at the unseen $15\times$ scale.

All experiments are implemented in PyTorch and conducted on a workstation equipped with a single NVIDIA GeForce RTX 5090 GPU. A CUDA-enabled environment supports model optimization and evaluation. The same hardware and software environment is maintained across SCOPE, baseline methods, ablation variants, and scale-transfer experiments to ensure consistent comparison.

\subsection{Ablation Study on Key Components}
\label{sec:ablation_study}

The ablation results in Table~\ref{tab:ablation_results} indicate that the full SCOPE network configuration provides the most balanced performance in both elevation accuracy and terrain-structure preservation. Here, Corr. denotes the slope-consistency score between reconstructed and reference terrain structures. Removing the bicubic base surface produces a clear deterioration, with RMSE increasing from 4.150 to 4.682 and Corr. decreasing from 0.659 to 0.521, showing that a low-frequency terrain reference is important for stable continuous reconstruction. Substituting bilinear or nearest interpolation for the base surface causes only moderate changes, but both remain slightly weaker than the default bicubic base surface. Replacing LAE with bilinear fusion also leads to a small yet consistent decline, suggesting that local reliability-aware fusion helps preserve fine-scale relief variation.

The refinement and loss-related variants mainly affect terrain morphology. Removing post-refinement markedly increases slope and aspect errors, indicating weaker control of local artifacts and high-frequency terrain details. The variant without the aspect-related loss shows the strongest degradation among the loss settings, especially in slope and aspect errors, which highlights the role of directional supervision in areas with complex local relief. Although the L1-only variant remains competitive in elevation accuracy, its weaker terrain-structure metrics show that elevation-wise fitting alone is insufficient for geomorphologically consistent DEM reconstruction.

The activation comparison further confirms the effectiveness of SASU as the default activation function in SCOPE. SiLU produces results very close to the full SASU-based model, suggesting that smooth gated nonlinearities are generally suitable for the elevation residual decoder. In contrast, GELU leads to a more evident degradation in both elevation accuracy and terrain-related metrics. Overall, the ablation study demonstrates that the bicubic base surface, adaptive local fusion, post-refinement operation, terrain-structure-aware supervision, and SASU activation jointly contribute to the final performance of the SCOPE network.

\begin{table*}[tbp]
\centering
\footnotesize
\setlength{\tabcolsep}{2pt}
\caption{Ablation study of the SCOPE network under the fixed $5\times$ reconstruction setting. Slope and Aspect denote slope-based and aspect-based reconstruction errors, respectively. Corr. denotes the slope-consistency score between reconstructed and reference terrain structures.}
\label{tab:ablation_results}
\par\vspace*{2pt}
\begin{tabular*}{\textwidth}{@{\extracolsep{\fill}} llrrrrrr @{}}
\toprule
\textbf{Category} & \textbf{Variant} & \textbf{RMSE $\downarrow$} & \textbf{MAE $\downarrow$} & \textbf{Slope $\downarrow$} & \textbf{Aspect $\downarrow$} & \textbf{PSNR $\uparrow$} & \textbf{Corr. $\uparrow$} \\
\midrule
Full model
& SCOPE network$^{*}$ & \textbf{4.150} & \textbf{2.816} & \textbf{12.051} & \textbf{46.473} & \textbf{77.48} & \textbf{0.659} \\
\midrule
\multirow{4}{*}{Base \& Fusion}
& No base & 4.682 & 3.238 & 14.718 & 57.076 & 73.88 & 0.521 \\
& Bilinear base & 4.153 & 2.817 & 12.051 & 46.571 & 77.43 & 0.659 \\
& Nearest base & 4.176 & 2.830 & 12.223 & 47.227 & 77.19 & 0.651 \\
& Bilinear fusion & 4.186 & 2.830 & 12.285 & 47.013 & 77.16 & 0.652 \\
\midrule
\multirow{3}{*}{Refinement \& Loss}
& No post-refinement & 4.885 & 3.372 & 17.468 & 63.935 & 73.64 & 0.436 \\
& L1 + L$_{\text{grad}}$ & 5.179 & 3.626 & 20.705 & 69.565 & 71.93 & 0.347 \\
& L1 only & 4.215 & 2.828 & 12.421 & 46.753 & 77.28 & 0.653 \\
\midrule
\multirow{2}{*}{Activation}
& GELU & 4.404 & 3.009 & 14.939 & 57.471 & 74.32 & 0.517 \\
& SiLU & 4.152 & 2.817 & 12.074 & 46.511 & 77.45 & 0.658 \\
\bottomrule
\end{tabular*}
\par\vspace*{2pt}
\begin{minipage}{\textwidth}
\footnotesize
\begin{enumerate}
\leftskip -1.5em
\item $\downarrow$ indicates that a lower value is better, while $\uparrow$ indicates that a higher value is better.
\item $^{*}$ denotes the full configuration comprising the bicubic base surface, the composite reconstruction--gradient--aspect loss, and the SASU activation function.
\end{enumerate}
\end{minipage}
\end{table*}

\subsection{Quantitative Comparison with Baselines}
\label{sec:quantitative_comparison}

The quantitative comparison in Table~\ref{tab:baseline_comparison} evaluates the SCOPE network against interpolation methods, fixed-scale super-resolution networks, and implicit neural reconstruction baselines under the fixed $5\times$ setting. Under this supervised scale, the SCOPE network ranks first across all six evaluation metrics. Relative to the strongest non-SCOPE result for each metric, RMSE, MAE, and slope error are reduced by 6.04\%, 5.28\%, and 7.16\%, respectively. Aspect error is reduced by 1.21\%, while PSNR and Corr. increase by 1.32 dB and 0.019. EBCF-CDEM is highly efficient with only 1.450M parameters and 11.153 GMACs. Compared with LIIF and LTE, SCOPE reduces the counted GMACs by 79.77\% and 77.62\%, respectively, while achieving the best overall reconstruction performance.

EDSR obtains the second-best RMSE, MAE, and slope error. Bicubic interpolation ranks second in aspect, PSNR, and Corr. SCOPE achieves the leading result across both elevation and terrain-structure metrics.

Among the INR-based methods, EBCF-CDEM obtains the lowest RMSE and MAE, showing competitive pointwise elevation fitting, but its slope, aspect, PSNR, and Corr. results are weaker than those of LIIF. LIIF is therefore used as the main INR-style visual baseline in the following qualitative comparison, while EBCF-CDEM remains the closest DEM-specific continuous reconstruction baseline in the quantitative analysis. Overall, the fixed-scale results indicate that the SCOPE network provides a more balanced reconstruction than interpolation, grid-based learning, and implicit neural representation baselines.

\begin{table*}[tbp]
\centering
\footnotesize
\setlength{\tabcolsep}{2pt}
\renewcommand{\arraystretch}{1.05}
\caption{Quantitative comparison with interpolation, fixed-scale super-resolution, and implicit neural reconstruction baselines under the fixed $5\times$ setting. All baseline architectures follow their official implementations or recommended model configurations and are trained and evaluated under the unified data and evaluation protocol. Params and GMACs are reported for batch size 1 with a $40\times40$ LR input and a $200\times200$ output, counting convolution, linear projection, attention matrix multiplication, and explicit basis-evaluation matrix multiplication. Red and blue indicate the best and second-best results for each metric, respectively. Lower RMSE, MAE, Slope, and Aspect indicate better performance, while higher PSNR and Corr. indicate better performance.}
\label{tab:baseline_comparison}
\begin{tabular*}{\textwidth}{@{\extracolsep{\fill}}llcccccccc@{}}
\toprule
\textbf{Category} & \textbf{Method} & \textbf{Params (M)} & \textbf{GMACs} & \textbf{RMSE $\downarrow$} & \textbf{MAE $\downarrow$} & \textbf{Slope $\downarrow$} & \textbf{Aspect $\downarrow$} & \textbf{PSNR $\uparrow$} & \textbf{Corr. $\uparrow$} \\
\midrule
\multirow{3}{*}{Interpolation} & Bicubic & 0 & -- & 4.733 & 3.205 & 13.304 & \second{47.041} & \second{76.16} & \second{0.640} \\
& Bilinear & 0 & -- & 5.551 & 3.747 & 14.172 & 49.389 & 74.81 & 0.618 \\
& Nearest & 0 & -- & 8.059 & 5.535 & 36.380 & 80.548 & 71.06 & 0.216 \\
\midrule
\multirow{4}{*}{Fixed-scale SR} & SRCNN & 0.479 & 19.150 & 7.844 & 3.890 & 13.858 & 50.629 & 67.75 & 0.576 \\
& EDSR & 21.582 & 34.577 & \second{4.417} & \second{2.973} & \second{12.981} & 49.803 & 75.36 & 0.612 \\
& SwinIR & 17.555 & 41.848 & 4.456 & 3.067 & 13.095 & 50.560 & 74.82 & 0.606 \\
& HAT & 5.287 & 10.214 & 5.742 & 3.230 & 14.842 & 56.412 & 71.61 & 0.520 \\
\midrule
\multirow{3}{*}{INR-based} & LIIF & 15.375 & 127.005 & 5.568 & 3.763 & 14.164 & 49.462 & 74.61 & 0.620 \\
& LTE & 15.958 & 114.827 & 5.807 & 4.077 & 14.198 & 51.125 & 71.94 & 0.596 \\
& EBCF-CDEM & 1.450 & 11.153 & 5.151 & 3.624 & 17.803 & 64.682 & 73.80 & 0.425 \\
\midrule
Ours & SCOPE network & 15.400 & 25.693 & \best{4.150} & \best{2.816} & \best{12.051} & \best{46.473} & \best{77.48} & \best{0.659} \\
\bottomrule
\end{tabular*}
\end{table*}

\subsection{Fixed-Scale Visual Analysis}
\label{sec:fixed_scale_visual_analysis}

The figures show representative land and marine examples, and quantitative metrics are computed over the complete predefined evaluation set. The qualitative comparisons include bicubic interpolation, the fixed-scale models EDSR and SwinIR, and the continuous reconstruction models LIIF and EBCF-CDEM. EBCF-CDEM provides the DEM-specific continuous reconstruction comparison.

\subsubsection{Visual Reconstruction Comparison}

\begin{figure}[tbp]
\centering
\includegraphics[width=0.9\linewidth]{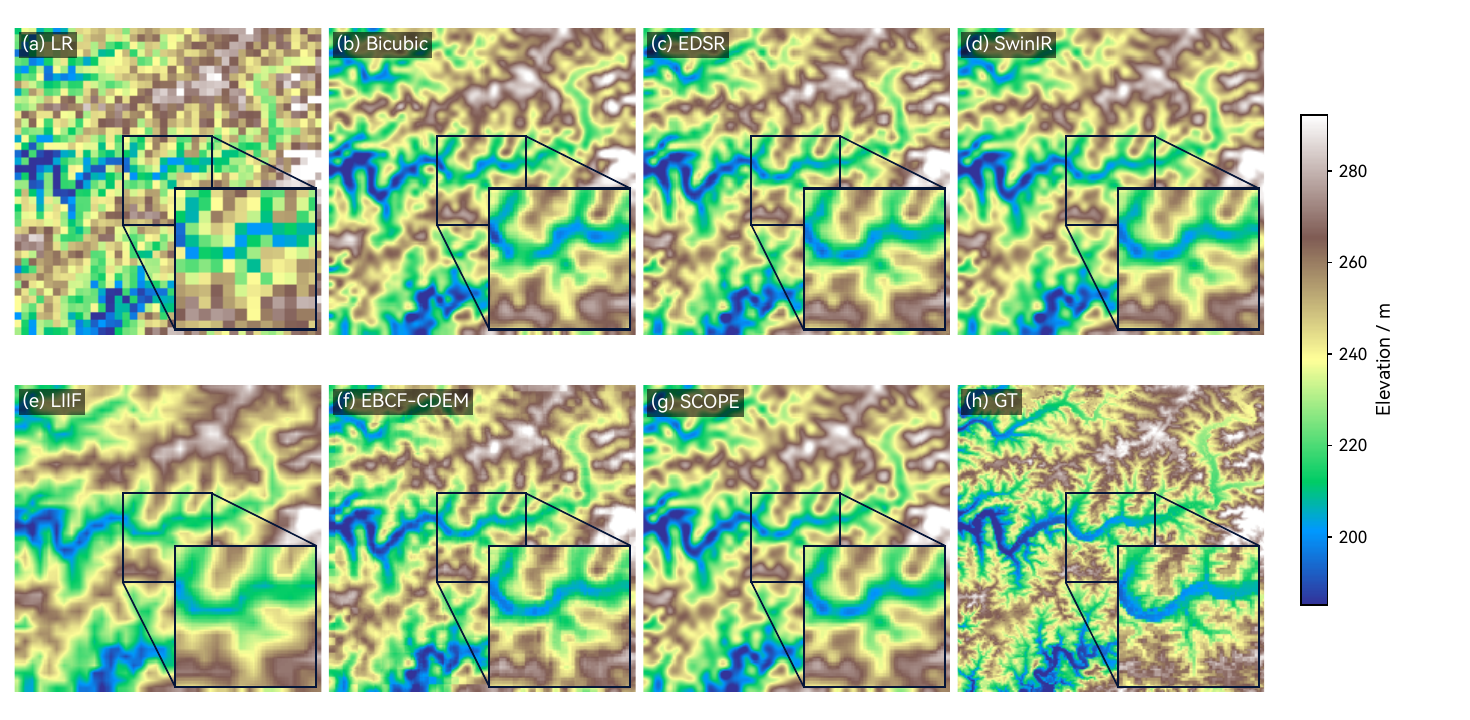}
\caption{Fixed-scale visual comparison on a representative land patch under the $5\times$ reconstruction setting. Panels (a)--(g) show the LR input, bicubic interpolation, EDSR, SwinIR, LIIF, EBCF-CDEM, and the SCOPE network, respectively, while panel (h) shows the GT HR DEM. All panels are compared over the same elevation range. The enlarged regions highlight local terrain details, where the SCOPE network better preserves ridge--valley structures and fine-scale relief variations.}
\label{fig:fixed5_reconstruction_land}
\end{figure}

\begin{figure}[tbp]
\centering
\includegraphics[width=0.9\linewidth]{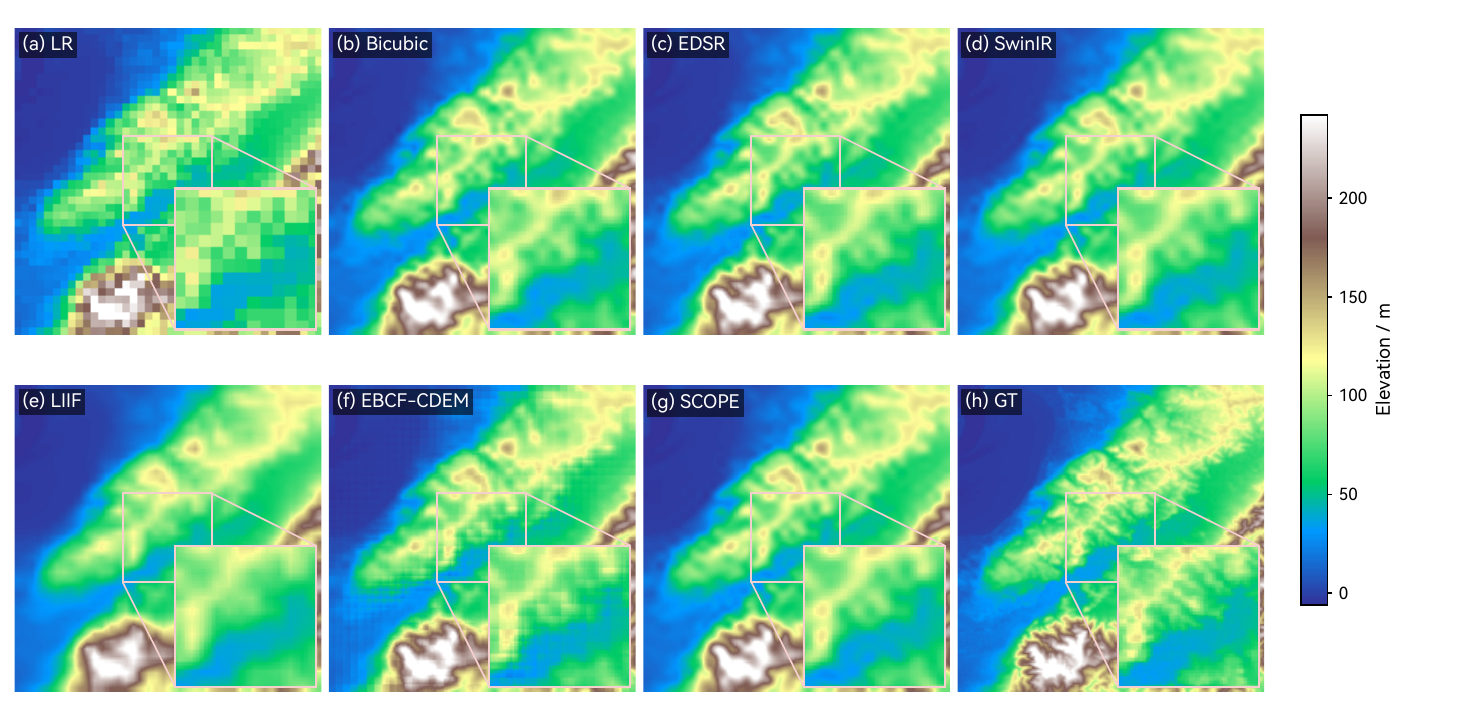}
\caption{Fixed-scale visual comparison on a representative marine and coastal patch under the $5\times$ reconstruction setting. Panels (a)--(g) show the LR input, bicubic interpolation, EDSR, SwinIR, LIIF, EBCF-CDEM, and the SCOPE network, respectively, while panel (h) shows the GT HR DEM. All panels are shown with a shared elevation range. The enlarged regions illustrate the reconstruction of coastal and bathymetric transition structures, where the SCOPE network produces a sharper and more coherent terrain surface.}
\label{fig:fixed5_reconstruction_marine}
\end{figure}

To further examine fixed-scale reconstruction behavior, representative land and marine patches are visualized under the supervised $5\times$ setting. As shown in Figs.~\ref{fig:fixed5_reconstruction_land} and~\ref{fig:fixed5_reconstruction_marine}, the comparison includes the HR reference, the LR input, bicubic interpolation, EDSR, SwinIR, LIIF, EBCF-CDEM, and the SCOPE network, with local enlargements highlighting fine-scale terrain details and geomorphological structures.

For the land patch, the LR input loses most narrow valley and ridge details, while bicubic interpolation restores the broad elevation pattern but produces an overly smooth surface. EDSR, SwinIR, and LIIF recover more coherent terrain than interpolation, yet small ridge--valley transitions and local elevation contrasts remain softened. For the marine and coastal patch, smooth bathymetric areas coexist with sharper nearshore transitions; here, LR degradation and bicubic interpolation weaken local gradients, and EDSR, SwinIR, and LIIF still blur subtle bathymetric and coastal relief variations. EBCF-CDEM remains important, but nearest-neighbor basing causes block artifacts in the marine example. In contrast, the SCOPE network better preserves valley continuity, ridge-like transitions, coastal gradients, and fine-scale relief variations, producing reconstructions that are visually closer to the HR reference. These observations are consistent with Table~\ref{tab:baseline_comparison}, where the SCOPE network achieves the best overall performance across elevation accuracy and terrain-related metrics.

\subsubsection{Elevation Error Distribution and Profile Analysis}

\begin{figure}[tbp]
\centering
\includegraphics[width=0.9\linewidth]{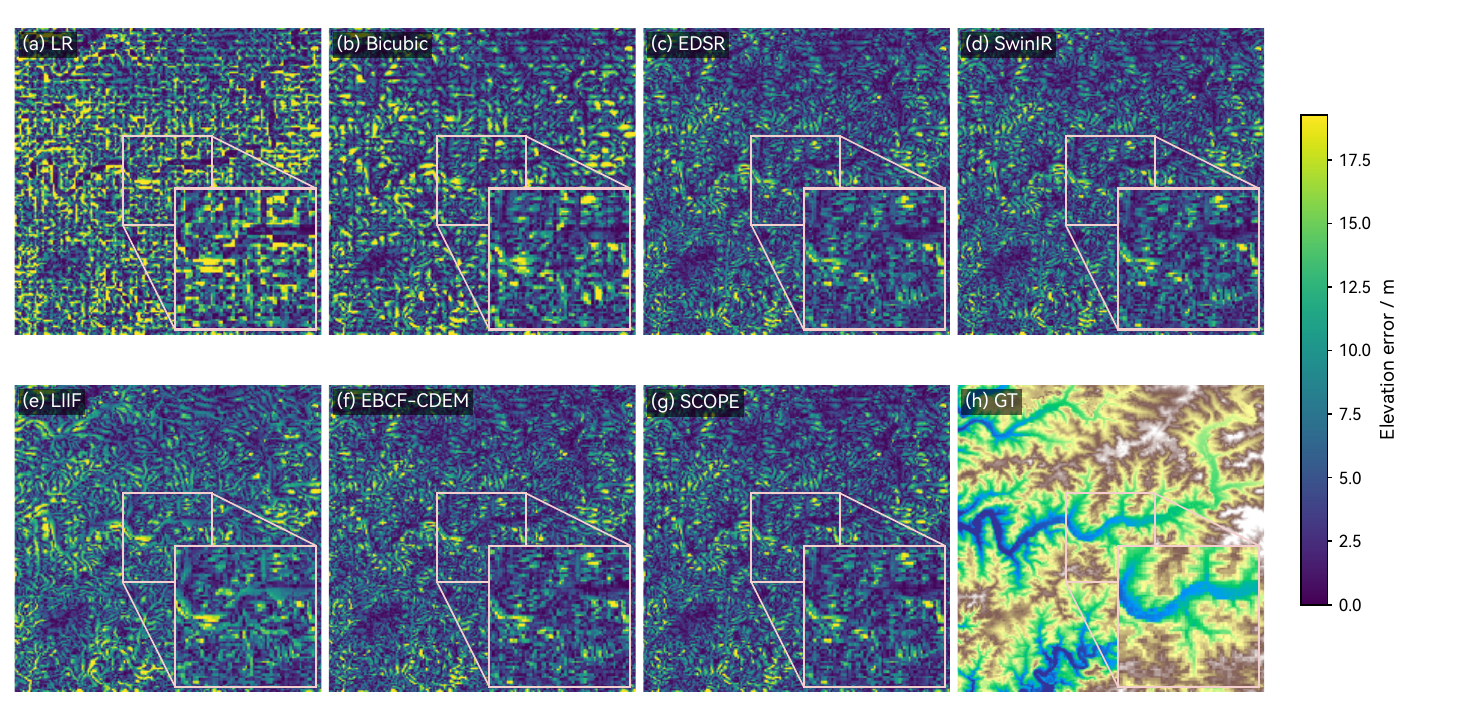}
\caption{Absolute elevation-error comparison on the representative land patch under the fixed $5\times$ reconstruction setting. Panels (a)--(g) show the absolute elevation errors of the LR input, bicubic interpolation, EDSR, SwinIR, LIIF, EBCF-CDEM, and the SCOPE network, respectively, with respect to the GT HR DEM shown in panel (h). The enlarged regions highlight error distributions around local ridge--valley structures.}
\label{fig:fixed5_error_land}
\end{figure}

\begin{figure}[tbp]
\centering
\includegraphics[width=0.9\linewidth]{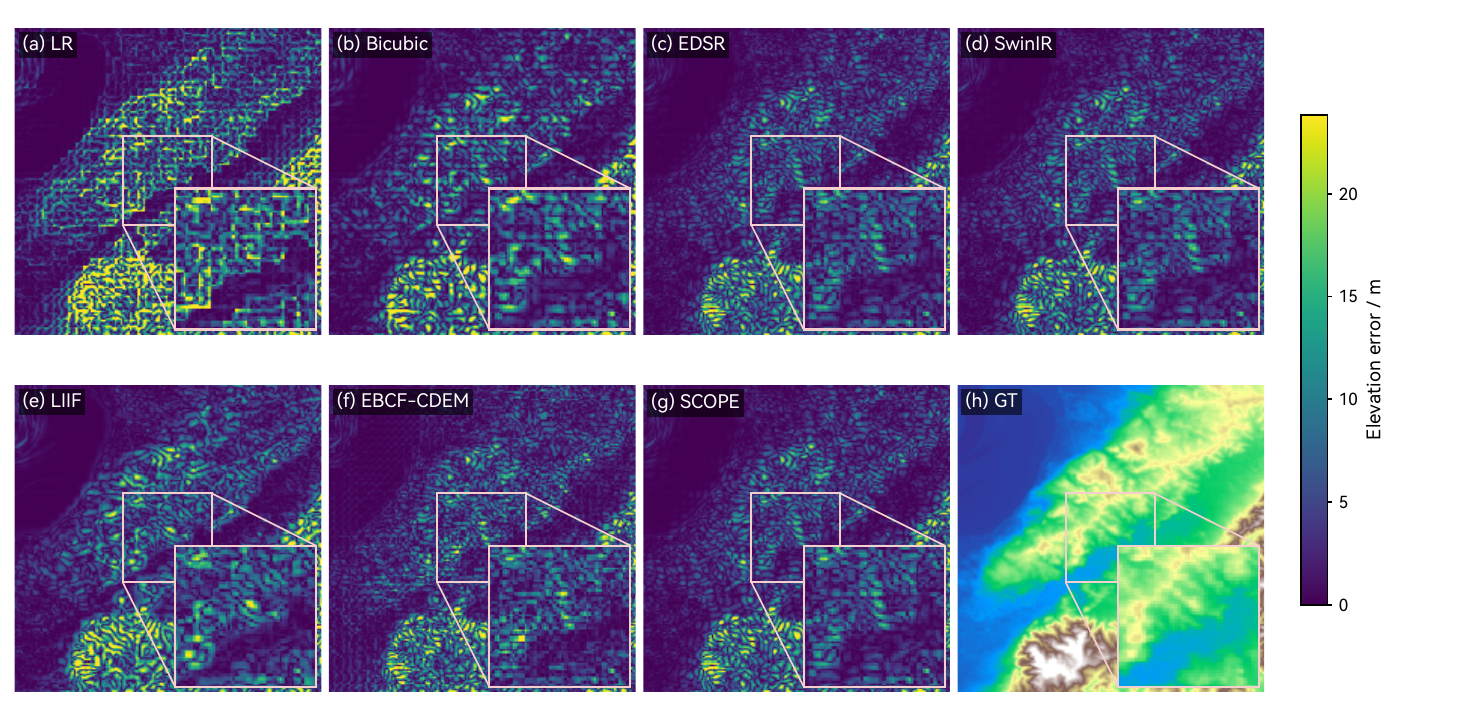}
\caption{Absolute elevation-error comparison on the representative marine and coastal patch under the fixed $5\times$ reconstruction setting. Panels (a)--(g) show the absolute elevation errors of the LR input, bicubic interpolation, EDSR, SwinIR, LIIF, EBCF-CDEM, and the SCOPE network, respectively, with respect to the GT HR DEM shown in panel (h). The enlarged regions emphasize errors near coastal and bathymetric transition zones.}
\label{fig:fixed5_error_marine}
\end{figure}

The absolute elevation-error maps and elevation profiles jointly reveal the spatial distribution and local magnitude of reconstruction errors. As shown in Figs.~\ref{fig:fixed5_error_land} and~\ref{fig:fixed5_error_marine}, the LR input produces strong block-related errors around narrow valleys, ridge boundaries, coastal slopes, and bathymetric transition zones. Bicubic interpolation reduces part of these artifacts but leaves evident errors along terrain boundaries, indicating that smooth interpolation cannot recover missing high-frequency relief structures.

Compared with bicubic interpolation, EDSR, SwinIR, and LIIF reduce part of the large-scale error but still show structured deviations where elevation changes rapidly. EBCF-CDEM is close to SCOPE, but SCOPE gives lower errors, especially along coastal transitions. In contrast, the SCOPE network exhibits a more spatially restrained error distribution, with fewer high-error responses in the enlarged regions. This indicates that SCOPE reduces not only average elevation deviation but also structured errors associated with complex terrain transitions.

\begin{figure}
\centering
\includegraphics[width=0.9\linewidth]{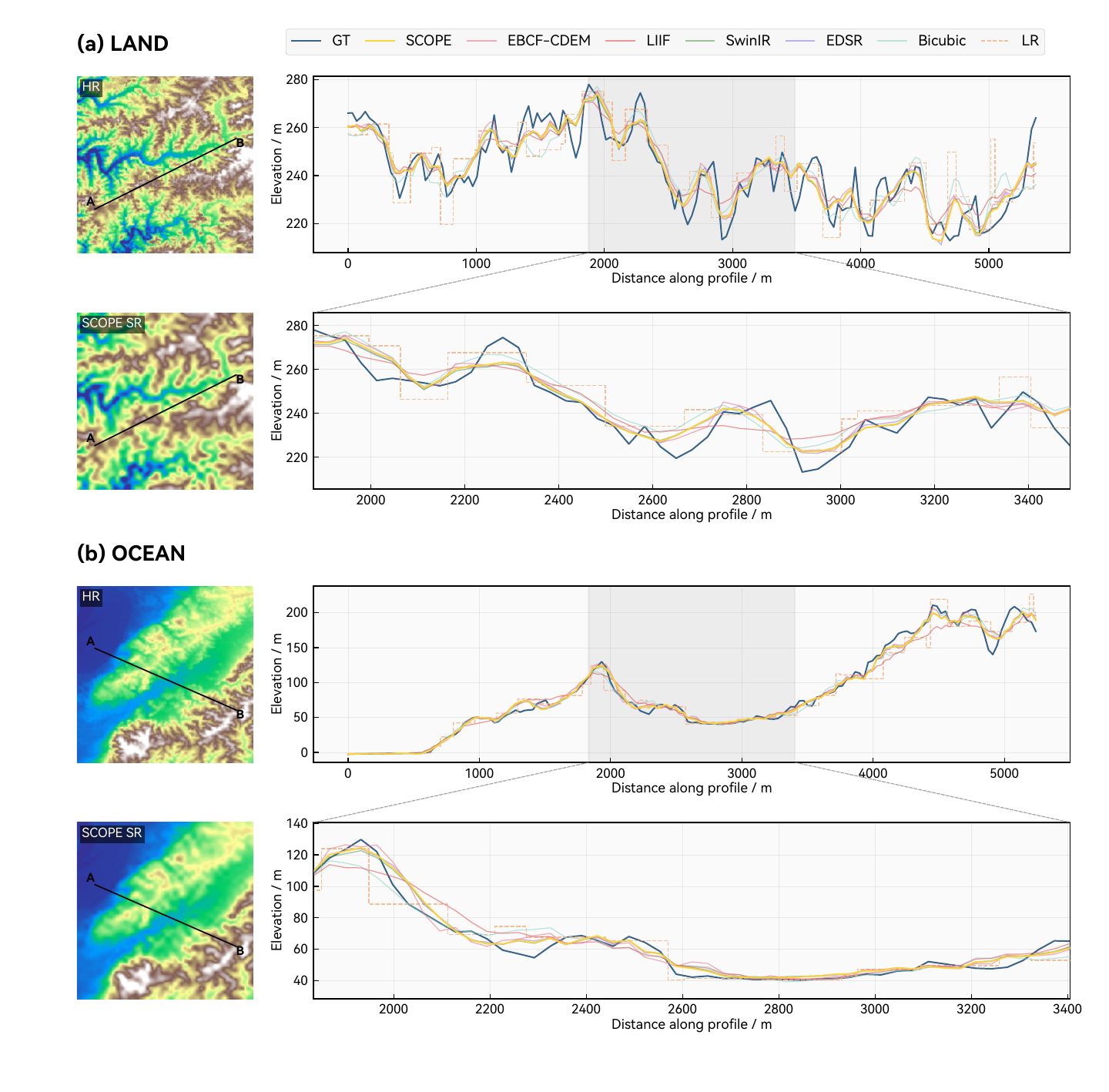}
\caption{Elevation profile comparison on representative land and ocean patches under the fixed $5\times$ reconstruction setting. For each case, the HR reference and SCOPE network reconstruction are shown with the same A--B profile line, while the full and zoomed elevation profiles compare HR, LR, bicubic interpolation, EDSR, SwinIR, LIIF, EBCF-CDEM, and the SCOPE network. The zoomed profiles provide a detailed view of local peak--valley reconstruction fidelity.}
\label{fig:fixed5_profile_comparison}
\end{figure}

The profile comparison in Fig.~\ref{fig:fixed5_profile_comparison} further assesses local elevation fidelity. All methods broadly follow the large-scale elevation trend, but differences remain around local peaks, valleys, and rapidly changing slopes. The LR profile shows step-like variations, bicubic interpolation smooths local extremes, and LIIF underestimates or slightly shifts several peak--valley amplitudes. EDSR and SwinIR follow the HR trajectory closely and remain strong fixed-scale elevation baselines, consistent with Table~\ref{tab:baseline_comparison}; however, their weaker aspect and Corr. metrics indicate that elevation fitting alone does not ensure terrain-structure consistency. EBCF-CDEM follows SCOPE closely, but larger local interpolation deviations remain. Compared with EDSR and SwinIR, the SCOPE network maintains similar or better profile fidelity while improving directional and geomorphological consistency under the fixed $5\times$ setting.

\subsubsection{Directional Terrain Structure Analysis}
\begin{figure}[htbp]
\centering
\includegraphics[width=0.9\linewidth]{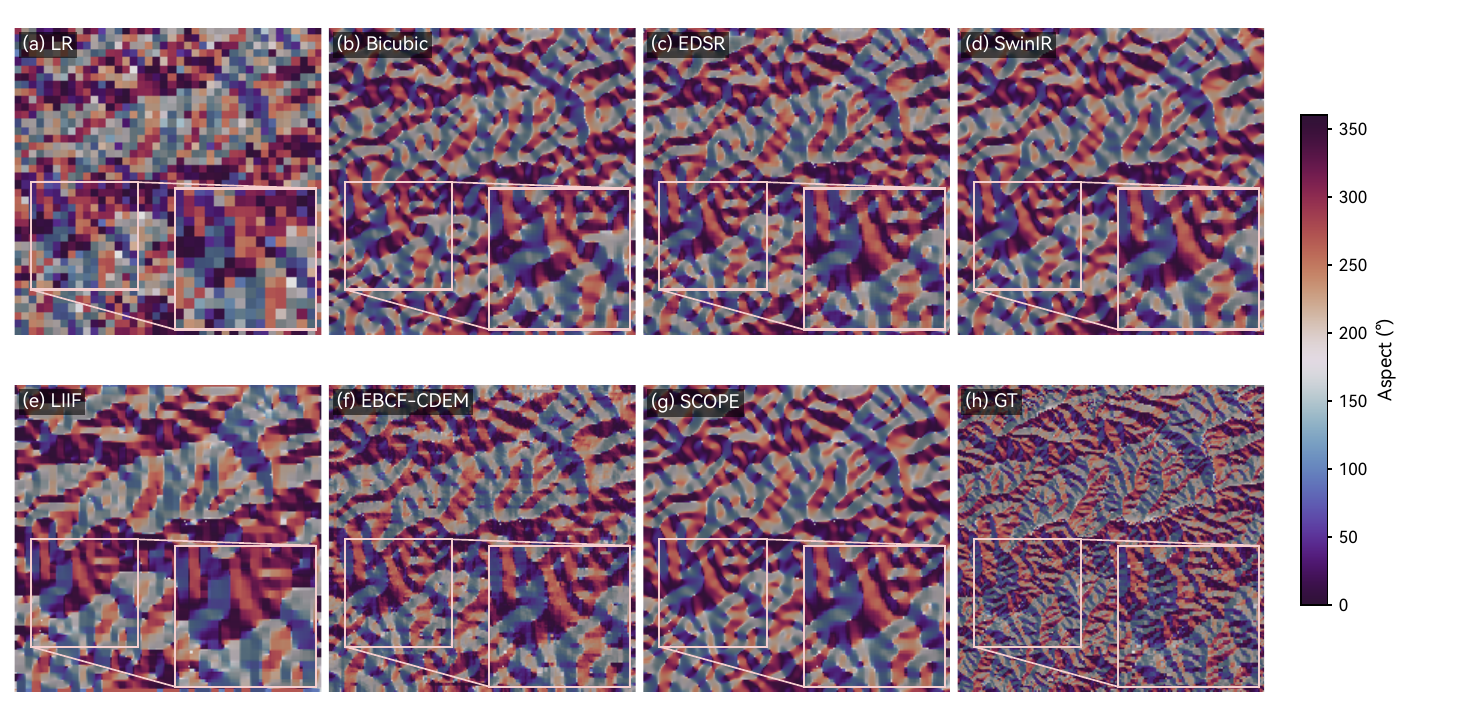}
\caption{Aspect-based directional terrain comparison on the representative land patch under the fixed $5\times$ reconstruction setting. The HR reference, LR input, bicubic interpolation, EDSR, SwinIR, LIIF, EBCF-CDEM, and the SCOPE network are visualized with the same cyclic aspect color mapping. The enlarged regions highlight local ridge--valley directional structures, where the SCOPE network better preserves fine-scale terrain orientation and directional continuity.}
\label{fig:fixed5_aspect_land}
\end{figure}

\begin{figure}[htbp]
\centering
\includegraphics[width=0.9\linewidth]{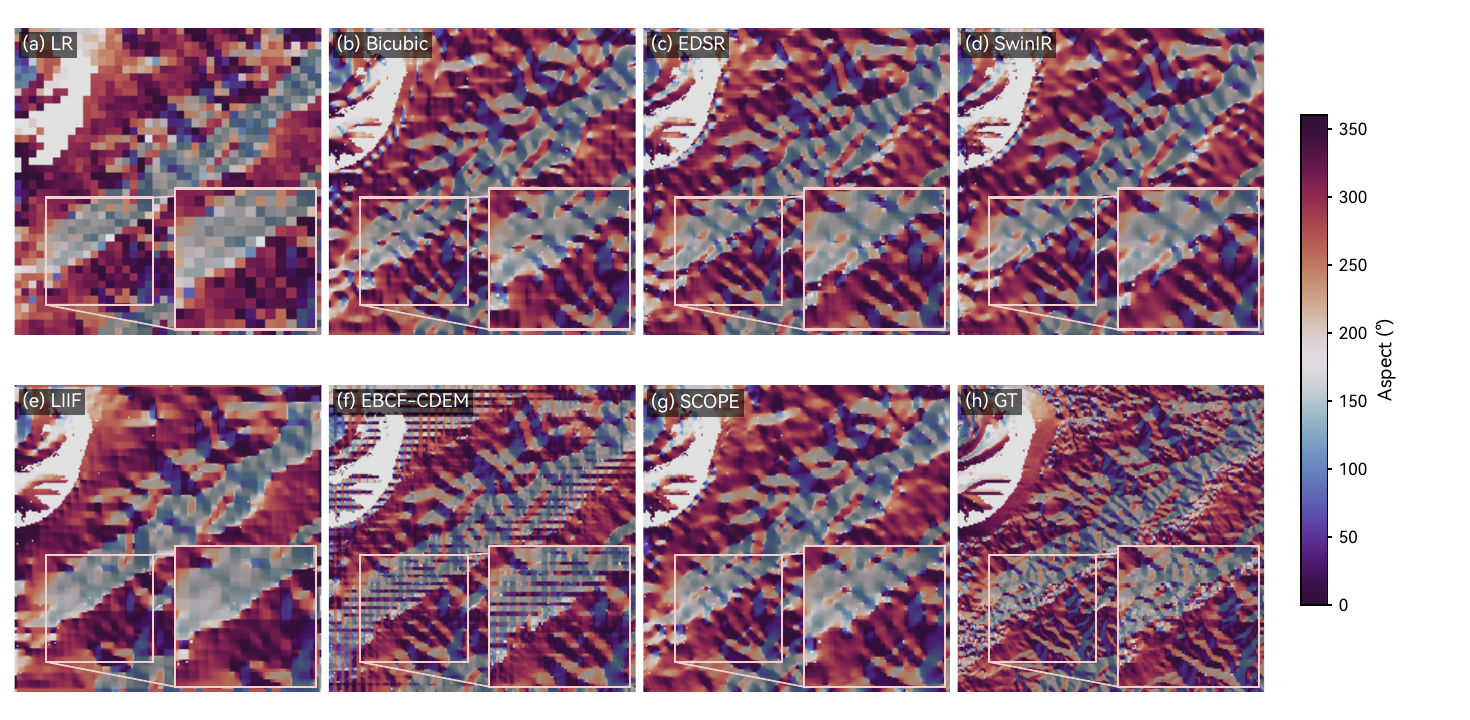}
\caption{Aspect-based directional terrain comparison on the representative marine and coastal patch under the fixed $5\times$ reconstruction setting. The HR reference, LR input, bicubic interpolation, EDSR, SwinIR, LIIF, EBCF-CDEM, and the SCOPE network are visualized with the same cyclic aspect color mapping. The enlarged regions emphasize coastal and bathymetric directional transitions, where the SCOPE network produces aspect patterns that are more consistent with the HR terrain structure.}
\label{fig:fixed5_aspect_marine}
\end{figure}

The aspect visualizations further examine whether the reconstructed DEMs preserve directionally consistent terrain structures. As shown in Figs.~\ref{fig:fixed5_aspect_land} and~\ref{fig:fixed5_aspect_marine}, the HR aspect maps exhibit organized directional patterns along ridge flanks, valley sides, coastal slopes, and bathymetric transition zones. These patterns are strongly degraded in the LR input, where coarse sampling produces block-like and locally inconsistent aspect responses. Bicubic interpolation partially restores directional continuity, but the recovered aspect fields remain over-smoothed, with weakened transitions around complex terrain boundaries.

EDSR, SwinIR, and LIIF recover plausible large-scale aspect patterns, but local artifacts remain: EDSR and SwinIR produce smoother, interpolation-like directional textures, whereas LIIF shows more regular block-like patterns in several enlarged areas. EBCF-CDEM preserves broad directions, but block-like aspect artifacts remain near complex transitions. In contrast, the SCOPE network better preserves fine-scale directional variations while maintaining spatial coherence. Its aspect fields are closer to the HR reference, with clearer local directional bands and fewer artificial discontinuities, supporting the quantitative improvement in the aspect-related metric and the preservation of local slope-facing direction.

\subsection{Cross-Domain Evaluation}
\label{sec:cross_dataset_region}

Cross-domain evaluation distinguishes regional transfer under a controlled degradation from transfer between different DEM products. The existing matched-pair experiment examines the effect of paired-product supervision, while the external marine evaluation tests the frozen main checkpoints on previously unused regions under both input constructions.

\subsubsection{Matched-Product Supervision and Transfer}

\begin{figure}[tbp]
\centering
\includegraphics[width=\linewidth]{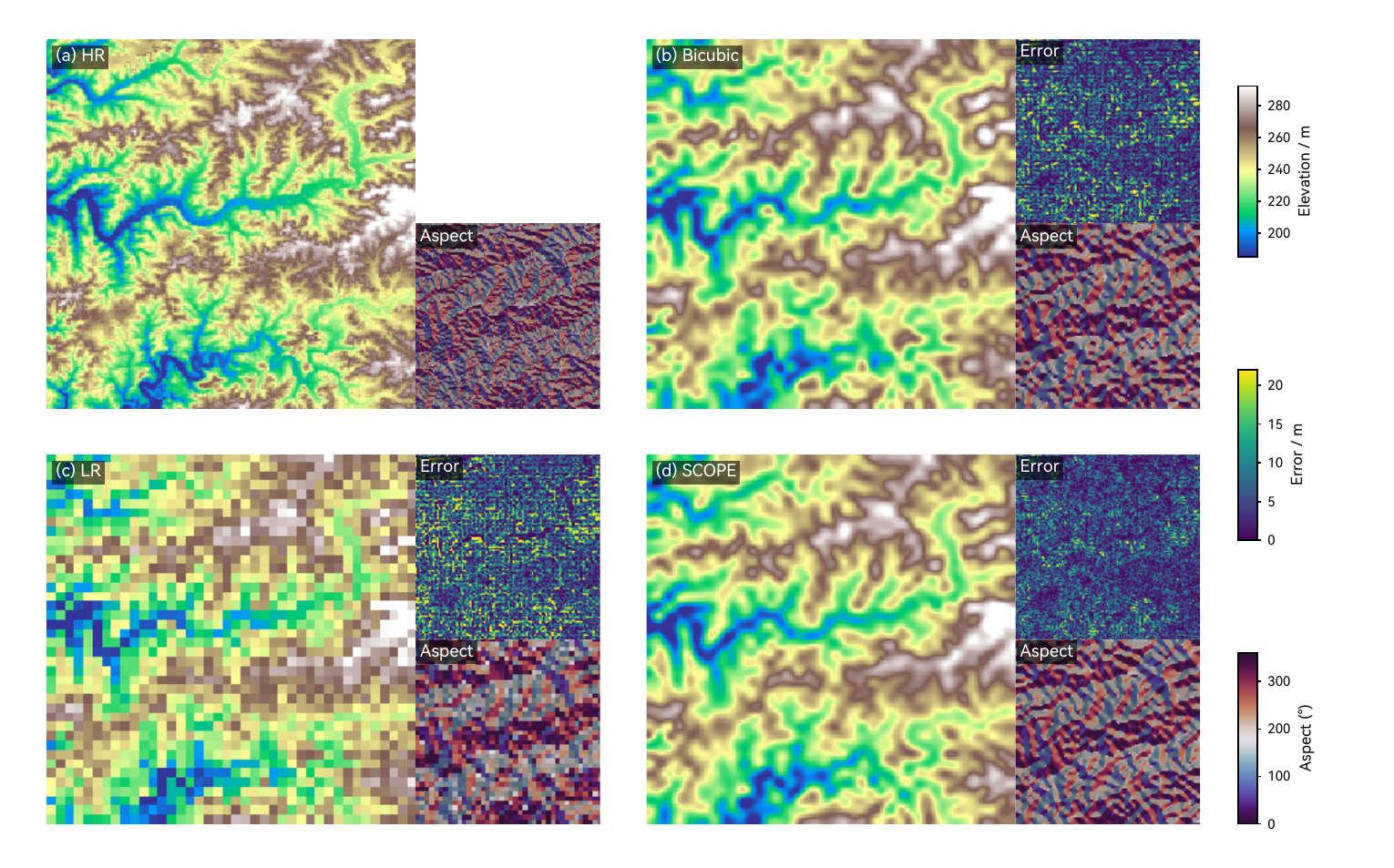}
\caption{Cross-dataset visual comparison on a representative land patch under the matched LR--HR paired setting. The HR reference, LR input, bicubic interpolation, and SCOPE network reconstruction are shown together with local error and aspect insets. Here, the SCOPE network corresponds to the directly trained SCOPE-Direct variant reported in Table~\ref{tab:cross_dataset_comparison}.}
\label{fig:paired_land_compact}
\end{figure}

\begin{figure}[tbp]
\centering
\includegraphics[width=\linewidth]{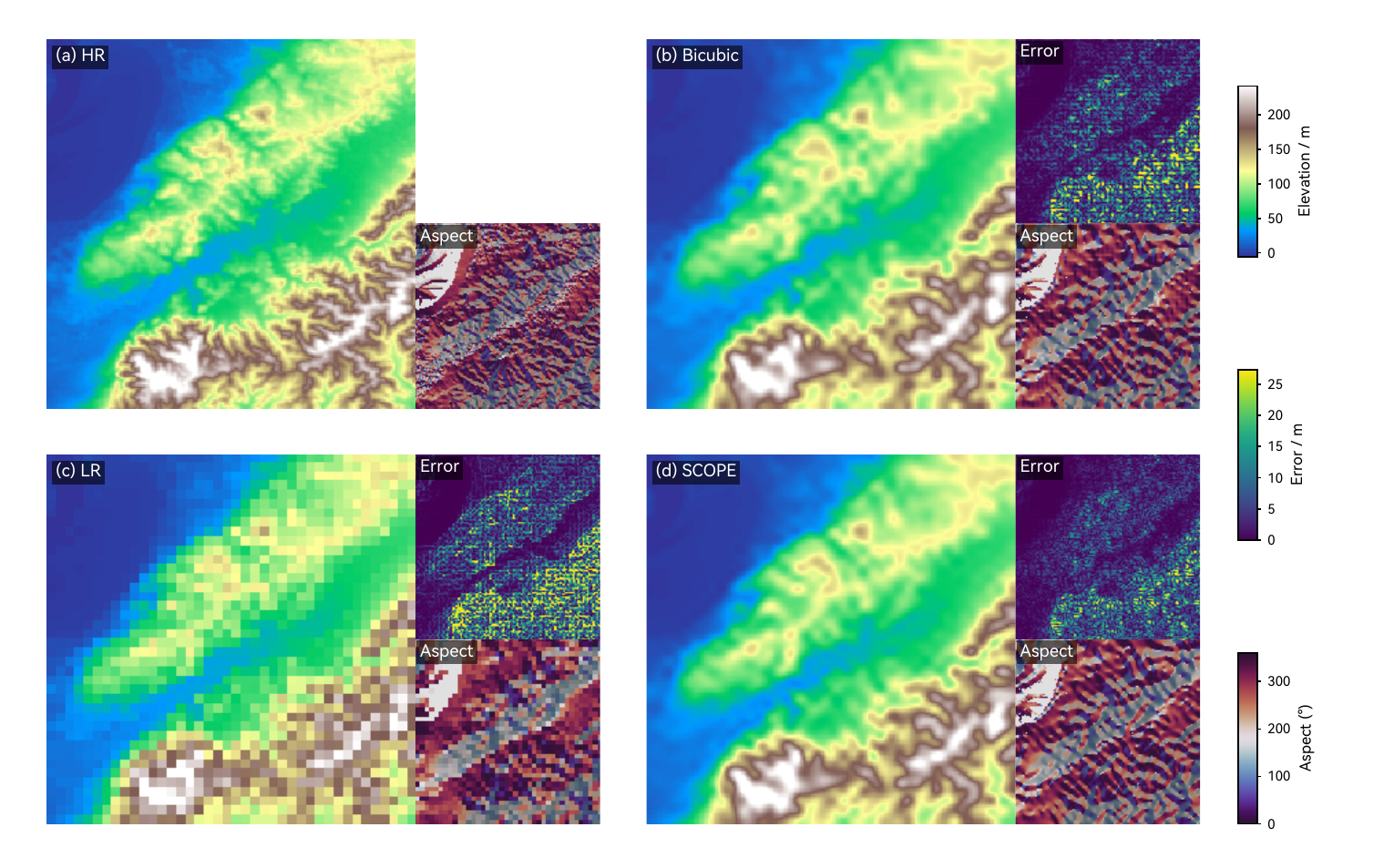}
\caption{Cross-dataset visual comparison on a representative marine and coastal patch under the matched LR--HR paired setting. The HR reference, LR input, bicubic interpolation, and SCOPE network reconstruction are shown together with local error and aspect insets. Here, the SCOPE network corresponds to the directly trained SCOPE-Direct variant reported in Table~\ref{tab:cross_dataset_comparison}.}
\label{fig:paired_marine_compact}
\end{figure}

Table~\ref{tab:cross_dataset_comparison} and Figs.~\ref{fig:paired_land_compact}--\ref{fig:paired_marine_compact} report the matched LR--HR experiment, in which the LR and HR DEMs are derived from different products. SCOPE-Direct and SCOPE-Transfer denote the paired-data and synthetic-data training variants in this experiment, respectively. The subsequent external marine experiment evaluates the separately identified frozen SCOPE-5F checkpoint.

\begin{table*}[htbp]
\centering
\footnotesize
\setlength{\tabcolsep}{2pt}
\renewcommand{\arraystretch}{1.05}
\caption{Matched-product supervision and transfer in the existing LR--HR paired experiment, distinct from the frozen external tests in Table~\ref{tab:external_cross_domain}. SCOPE-Direct is trained directly on matched paired data, while SCOPE-Transfer is trained with synthetic downsampling and evaluated on the paired data. Red and blue indicate the best and second-best results for each metric, respectively. Lower RMSE, MAE, Slope, and Aspect indicate better performance, while higher PSNR and Corr. indicate better performance.}
\label{tab:cross_dataset_comparison}
\begin{tabular*}{\textwidth}{@{\extracolsep{\fill}}lcccccc@{}}
\toprule
\textbf{Method} & \textbf{RMSE $\downarrow$} & \textbf{MAE $\downarrow$} & \textbf{Slope $\downarrow$} & \textbf{Aspect $\downarrow$} & \textbf{PSNR $\uparrow$} & \textbf{Corr. $\uparrow$} \\
\midrule
Bicubic & \second{8.452} & \second{6.145} & 15.230 & \best{57.081} & \best{69.53} & \best{0.537} \\
Bilinear & 8.989 & 6.544 & 15.931 & \second{57.565} & \second{69.17} & \second{0.528} \\
Nearest & 10.774 & 7.733 & 37.668 & 83.901 & 67.16 & 0.172 \\
\midrule
SCOPE-Direct & \best{8.164} & \best{5.880} & \second{15.184} & 60.453 & 69.05 & 0.507 \\
SCOPE-Transfer & 8.546 & 6.223 & \best{14.963} & 61.152 & 68.78 & 0.496 \\
\bottomrule
\end{tabular*}
\end{table*}

SCOPE-Direct achieves the lowest RMSE and MAE, while SCOPE-Transfer obtains the lowest grid-space slope error. Bicubic interpolation gives the lowest aspect error and the highest PSNR and Corr. The RMSE and MAE of SCOPE-Transfer are 8.546 and 6.223, compared with 8.452 and 6.145 for bicubic. Figures~\ref{fig:paired_land_compact} and~\ref{fig:paired_marine_compact} show the corresponding land and marine reconstructions for SCOPE-Direct.

\subsubsection{Frozen-Model Evaluation on External Marine Regions}
\label{sec:external_results}

Table~\ref{tab:external_cross_domain} reports the same six metrics used in the other comparisons for the two input settings on the same external reference patches. Under self-downsampling, SCOPE-5F achieves the lowest RMSE, MAE, Slope, and Aspect and the highest PSNR and Corr. among all compared methods, including bicubic interpolation. Its RMSE of 3.4009~m and MAE of 2.1734~m correspond to reductions of 10.11\% and 8.97\% relative to bicubic, and 2.88\% and 2.73\% relative to EDSR. Relative to the DEM-specific EBCF-CDEM baseline, its RMSE is reduced by 19.40\%. The accompanying improvement in terrain-related metrics supports cross-region generalization under the controlled degradation used to form the LR inputs.

\begin{table*}[htbp]
\centering
\footnotesize
\setlength{\tabcolsep}{2pt}
\renewcommand{\arraystretch}{1.10}
\caption{Frozen-model external marine evaluation at $5\times$ (nominal $450\rightarrow90$~m), using the same 12 reference rasters and 3,825 valid patches per method in both input settings. All metrics are averaged over patches. Slope, Aspect, and Corr. follow the definitions used throughout the paper and correspond to RMSE-Slope, RMSE-Aspect, and SlopeCorr in the evaluation logs. Red and blue indicate the best and second-best distinct values within each setting, including ties at the displayed precision.}
\label{tab:external_cross_domain}
\begin{tabular*}{\textwidth}{@{\extracolsep{\fill}}llrrrrrr@{}}
\toprule
\textbf{Input setting} & \textbf{Method} & \textbf{RMSE $\downarrow$} & \textbf{MAE $\downarrow$} & \textbf{Slope $\downarrow$} & \textbf{Aspect $\downarrow$} & \textbf{PSNR $\uparrow$} & \textbf{Corr. $\uparrow$} \\
\midrule
\multirow{5}{*}{Self-downsampled} & Bicubic & 3.7834 & 2.3876 & 15.0834 & \second{51.3582} & 77.4655 & \second{0.6045} \\
 & LIIF-MS & 73.6168 & 73.3150 & 15.7968 & 53.3109 & 44.5991 & 0.5898 \\
 & EBCF-CDEM & 4.2196 & 2.8100 & 20.5145 & 71.4522 & 73.6952 & 0.3776 \\
 & EDSR & \second{3.5019} & \second{2.2343} & \second{14.1920} & 52.4866 & \second{77.5683} & 0.5953 \\
 & SCOPE-5F & \best{3.4009} & \best{2.1734} & \best{13.9307} & \best{51.0133} & \best{78.3732} & \best{0.6241} \\
\midrule
\multirow{5}{*}{Matched GEBCO\_2023} & Bicubic & \best{9.0587} & \best{5.7358} & \best{17.6567} & \second{68.0274} & \best{68.6858} & \second{0.4574} \\
 & LIIF-MS & 75.2739 & 74.2693 & 18.0105 & \best{66.8491} & 44.4350 & \best{0.4586} \\
 & EBCF-CDEM & 9.5928 & 6.0837 & 21.9579 & 79.8568 & 67.6670 & 0.3199 \\
 & EDSR & 9.4070 & 5.9258 & 18.0412 & 73.0962 & 68.1560 & 0.4157 \\
 & SCOPE-5F & \second{9.3932} & \second{5.9206} & \second{17.9543} & 72.5997 & \second{68.1863} & 0.4230 \\
\bottomrule
\end{tabular*}
\end{table*}

With matched GEBCO\_2023 inputs, SCOPE-5F leads the learned methods in RMSE, MAE, Slope, and PSNR. Its RMSE is 2.08\% lower than EBCF-CDEM, and its RMSE and MAE are lower than EDSR by 0.0138~m and 0.0052~m, respectively. Bicubic achieves the lowest RMSE, MAE, and Slope and the highest PSNR across all methods; SCOPE-5F's RMSE and MAE are 3.69\% and 3.22\% higher than bicubic. LIIF-MS attains the lowest Aspect error and the highest Corr., with RMSE and MAE of 75.2739~m and 74.2693~m.

\subsection{Reconstruction Beyond the Supervised Scale}
\label{sec:scale_transfer}

The central scale-transfer experiment evaluates whether a terrain representation learned solely under fixed $5\times$ supervision remains effective when directly queried at the unseen $15\times$ scale. Within the same spatial support, this setting increases the output-grid density by a factor of nine, while SCOPE-5F receives neither $15\times$ target supervision nor target-scale retraining. It therefore provides a stricter test than arbitrary-scale querying within or near the supervised scale range.

As reported in Table~\ref{tab:scale_transfer_15x}, SCOPE-5F achieves an RMSE of 6.753 and an MAE of 4.609. Relative to LIIF-MS, these errors are reduced by 21.66\% and 22.00\%, respectively, while the reductions over EBCF-CDEM reach 34.21\% and 36.12\% within the same unseen-$15\times$ land evaluation subset. SCOPE-5F also improves RMSE, MAE, slope error, PSNR, and Corr. over bicubic interpolation. Aspect errors are 71.534 for SCOPE-5F and 71.386 for bicubic.

The controlled SCOPE variants show closely matched performance. SCOPE-5F is numerically better than SCOPE-15D across all six metrics in this evaluation. SCOPE-15FT produces the best overall result, with RMSE 0.08\% lower than SCOPE-5F. SCOPE-5MS yields an RMSE of 6.769 and an MAE of 4.613, compared with 6.753 and 4.609 for SCOPE-5F.

\begin{figure*}[htbp]
\centering
\includegraphics[width=0.95\textwidth]{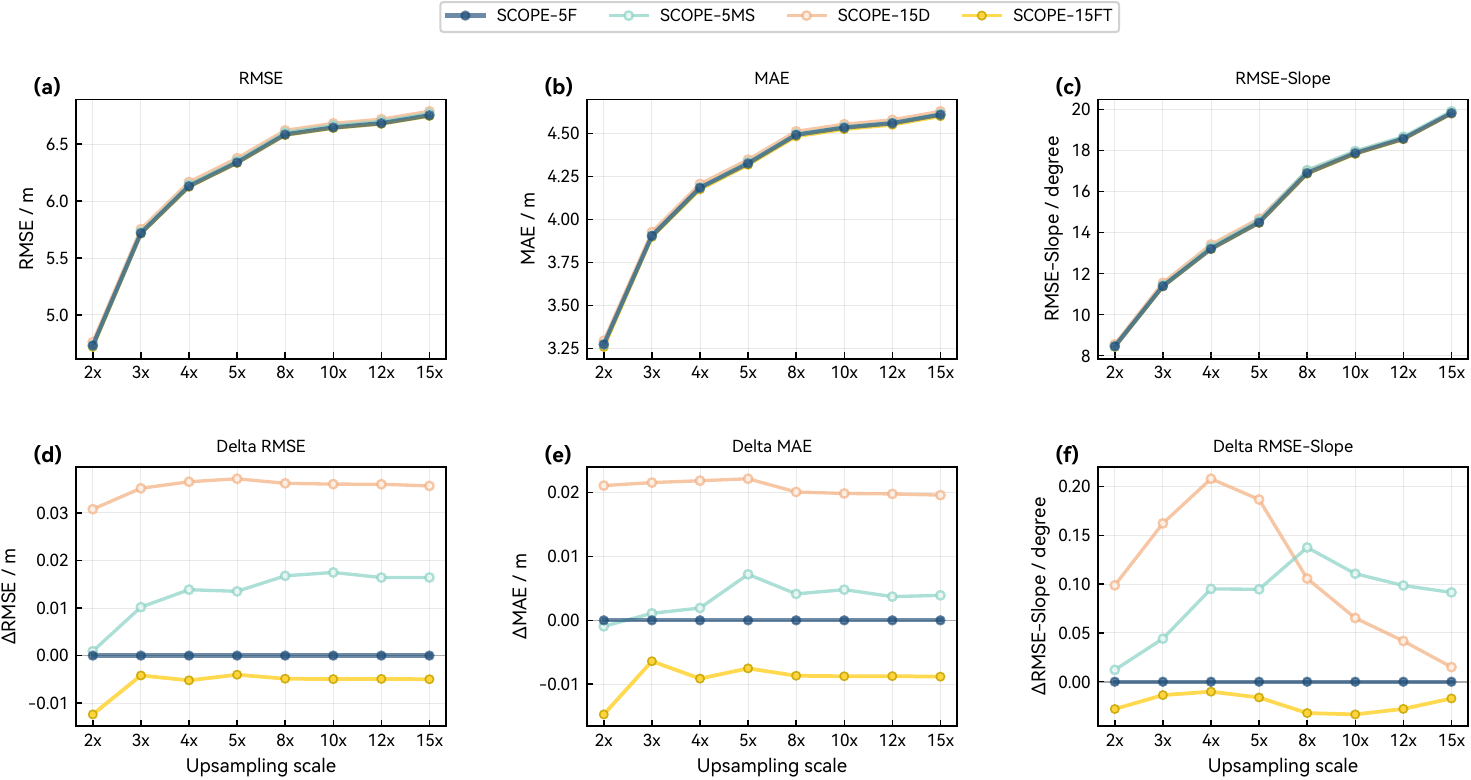}
\caption{Scale-transfer behavior of different SCOPE network training protocols from $2\times$ to $15\times$. The top row reports absolute RMSE, MAE, and slope error, while the bottom row reports metric differences computed as each compared protocol minus SCOPE-5F at the same scale. SCOPE-5F denotes fixed $5\times$ training, SCOPE-5MS denotes multiscale training, SCOPE-15D denotes direct $15\times$ training, and SCOPE-15FT denotes $5\times$ training followed by $15\times$ fine-tuning.}
\label{fig:scale_sweep_metrics}
\end{figure*}

Figure~\ref{fig:scale_sweep_metrics} reports the absolute errors and differences relative to SCOPE-5F across $2\times$--$15\times$. The fixed-$5\times$ model achieves the reported $15\times$ reconstruction directly from its learned coefficient field. EBCF-CDEM uses 1.450M parameters and 84.799 GMACs at $15\times$. Compared with LIIF-MS and EBCF-CDEM, SCOPE-5F reduces the counted GMACs by 97.23\% and 69.09\%, respectively, while achieving lower RMSE, MAE, slope error, and higher Corr.

\begin{table*}[htbp]
\centering
\footnotesize
\setlength{\tabcolsep}{2pt}
\renewcommand{\arraystretch}{1.05}
\caption{Quantitative comparison at the target $15\times$ reconstruction scale. LIIF-MS is trained with $5\times$ as the primary scale and random downsampling across $\{2\times,3\times,4\times\}$ for multi-scale supervision. SCOPE-5F denotes fixed $5\times$ training only, SCOPE-5MS denotes multiscale training up to $5\times$, SCOPE-15D denotes direct $15\times$ training, and SCOPE-15FT denotes $5\times$ training followed by $15\times$ fine-tuning. Params and GMACs are reported for batch size 1 with a $40\times40$ LR input and a $600\times600$ output, counting convolution, linear projection, attention matrix multiplication, and explicit basis-evaluation matrix multiplication. Red and blue indicate the best and second-best results for each metric, respectively. Lower RMSE, MAE, Slope, and Aspect indicate better performance, while higher PSNR and Corr. indicate better performance.}
\label{tab:scale_transfer_15x}
\begin{tabular*}{\textwidth}{@{\extracolsep{\fill}}lcccccccc@{}}
\toprule
\textbf{Method} & \textbf{Params (M)} & \textbf{GMACs} & \textbf{RMSE $\downarrow$} & \textbf{MAE $\downarrow$} & \textbf{Slope $\downarrow$} & \textbf{Aspect $\downarrow$} & \textbf{PSNR $\uparrow$} & \textbf{Corr. $\uparrow$} \\
\midrule
Bicubic & 0 & -- & 7.657 & 5.212 & 21.468 & \best{71.386} & 69.57 & 0.447 \\
LIIF-MS & 15.375 & 946.533 & 8.620 & 5.909 & 22.293 & 73.185 & 68.61 & 0.426 \\
EBCF-CDEM & 1.450 & 84.799 & 10.265 & 7.215 & 29.331 & 96.334 & 65.01 & 0.182 \\
\midrule
SCOPE-5F & 15.400 & 26.208 & \second{6.753} & \second{4.609} & \second{19.824} & 71.534 & \best{70.45} & \second{0.455} \\
SCOPE-5MS & 15.400 & 26.208 & 6.769 & 4.613 & 19.915 & 71.667 & 70.43 & 0.454 \\
SCOPE-15D & 15.400 & 26.208 & 6.789 & 4.629 & 19.839 & 72.028 & 70.37 & 0.446 \\
SCOPE-15FT & 15.400 & 26.208 & \best{6.748} & \best{4.600} & \best{19.807} & \second{71.492} & \best{70.45} & \best{0.456} \\
\bottomrule
\end{tabular*}
\end{table*}

\section{Discussion}
\label{sec:discussion}
\subsection{Crossing the Supervision Boundary: From Known Scales to Unseen Reconstruction}
\label{sec:discussion_scale_boundary}

The reconstruction problem considered here combines a supervision constraint with a computational requirement: labels are available at a coarser output resolution, while deployment may request a denser terrain grid. SCOPE addresses this combination through reusable coefficient-field prediction rather than through continuous coordinate querying alone. The fixed-$5\times$ to unseen-$15\times$ experiment evaluates this design outside the supervised output scale, increasing the query density ninefold without additional target-scale training labels.

SCOPE supports this transition by separating terrain representation learning from output-grid construction. A latent coefficient field is learned on the LR feature grid and parameterizes reusable local elevation-residual functions. Changing the reconstruction factor therefore does not require a new scale-specific output head or a different terrain representation; only the density at which the coefficient field is evaluated is altered. The bicubic base preserves the broad elevation trend, while local basis evaluation and LAE reconstruct terrain variations from neighboring coefficient vectors. This design makes the reconstructed surface depend on a reusable terrain field rather than on isolated query-wise elevation answers.

The land scale-transfer results support the usefulness of the learned representation beyond its supervised resolution. With the input held at nominal 450~m spacing, the model trained against 90~m labels remains effective when queried at 30~m. Direct target-scale training and fine-tuning yield results close to those of the fixed-$5\times$ model. The evidence therefore supports useful reconstruction with supervision at the coarser output resolution, without requiring a claim that fixed-scale training is intrinsically superior to the alternative protocols.

\subsection{Implications for Earth Observation Terrain Products}

The coefficient-field formulation allows a terrain representation learned from available paired data to be evaluated on different output grids. For applications requiring finer sampling than the training references provide, this can reduce dependence on separate scale-specific models and target-resolution training labels. The land results demonstrate a useful case: a model trained with nominal 90~m supervision improves on interpolation when evaluated against 30~m references. The deployment value lies in flexible reconstruction from existing products, rather than in replacing new terrain observations.

The external marine tests clarify a separate requirement for such use. Under self-downsampling, SCOPE retains an accuracy advantage on the held-out regions, whereas matched GEBCO inputs do not yield a gain over bicubic. Its cross-product elevation errors are close to EDSR, and LIIF-MS leads the Aspect and Corr. metrics despite its large elevation errors. The earlier matched-product experiment also distributes the best metrics across SCOPE-Direct, SCOPE-Transfer, and bicubic. These metric-specific rankings do not establish a uniform cross-product advantage. Regional generalization under a common degradation model and transfer between real products are therefore different capabilities. Differences in acquisition source, vertical reference, interpolation history, and terrain distribution can affect matched-product reconstruction; the aggregate results do not isolate their individual contributions.

\subsection{Why Lightweight Local Attentive Ensemble Is Sufficient}

Table~\ref{tab:local_fusion_discussion} shows that LAE provides small but consistent improvements over Transformer-style fusion across the terrain-related metrics. At each query coordinate $\mathbf{q}$, the fusion module receives only four local residual candidates $r_i(\mathbf{q})$ together with their relative displacements $\Delta\mathbf{q}_i$. Its role is therefore not to learn long-range interactions or rich token dependencies, but to estimate compatibility weights for a compact set of geometry-related residual candidates.

Transformer-style fusion remains a powerful general interaction mechanism, particularly for feature extraction and larger token sets, as already exploited by the upstream encoder. At the query stage, however, most terrain context has been encoded into the latent coefficient field, leaving only a low-dimensional local aggregation problem. LAE introduces a more direct inductive bias by mapping local residual candidates and geometric offsets to adaptive compatibility weights. Its slightly lower elevation and terrain-structure errors indicate that this compact weighting mechanism is sufficient for local basis evaluation, while avoiding unnecessary query-side complexity.

\begin{table}[htbp]
\centering
\footnotesize
\setlength{\tabcolsep}{2pt}
\renewcommand{\arraystretch}{1.05}
\caption{Comparison between MLP-based LAE and Transformer-style local fusion. Lower values indicate better performance.}
\label{tab:local_fusion_discussion}
\begin{tabular}{lcccc}
\toprule
\textbf{Fusion} & \textbf{RMSE} & \textbf{MAE} & \textbf{Slope} & \textbf{Aspect} \\
\midrule
MLP-based LAE & \best{4.150} & \best{2.816} & \best{12.051} & \best{46.473} \\
Transformer-style fusion & 4.186 & 2.832 & 12.302 & 47.061 \\
\bottomrule
\end{tabular}
\end{table}

\subsection{Effect of Sign-Adaptive Activation}

The activation ablation shows lower errors with SASU than with GELU in the reported run, reducing RMSE from 4.404 to 4.150 and grid-space slope error from 14.939 to 12.051. SASU and SiLU produce nearly identical results. SASU is retained as the default activation, while the comparison supports the use of smooth gated nonlinearities without establishing a substantial advantage over SiLU.

SASU applies different smooth nonlinear responses to non-negative and negative activations without introducing additional parameters, providing a lightweight component for residual decoding.

\subsection{Computational Scaling and Inference Efficiency}

The Params and GMACs reported in Tables~\ref{tab:baseline_comparison} and~\ref{tab:scale_transfer_15x} show a distinct computational response to denser querying. When the reconstruction scale increases from $5\times$ to $15\times$, the counted GMACs increase by only 0.515 for SCOPE, compared with 819.528 for LIIF and 73.646 for EBCF-CDEM. This contrast arises from what is repeated during HR querying. Query-wise implicit baselines repeatedly apply linear layers in an MLP to dense HR coordinates, whereas SCOPE generates a latent coefficient field once from spatially organized LR feature maps. The query coordinate then mainly triggers coefficient sampling, basis-function combination, and lightweight LAE weighting, rather than another high-dimensional coordinate-to-elevation regression. This computational distinction can be summarized as

\begin{equation}
\begin{aligned}
\mathrm{MAC}_{\mathrm{query}}(Q)
&=
C_{\mathrm{enc}}
+
Q\,C_{\mathrm{query}},\\
\mathrm{MAC}_{\mathrm{SCOPE}}(Q)
&=
C_{\mathrm{field}}
+
Q\,C_{\mathrm{eval}}.
\end{aligned}
\label{eq:scope_complexity_discussion}
\end{equation}

where $Q$ is the number of output query points, $C_{\mathrm{query}}$ denotes the per-query MLP cost in query-wise implicit baselines, and $C_{\mathrm{field}}$ includes LR-side feature extraction and coefficient prediction. The effective per-output cost $C_{\mathrm{eval}}$ includes the counted operations of local field evaluation, LAE fusion, and output-grid post-refinement. Under the evaluated configurations, this cost is much smaller than repeated high-dimensional query prediction. Both formulations retain a term linear in $Q$; the advantage is a lower marginal arithmetic cost, not constant-cost dense reconstruction. The LR-side work dominates SCOPE's reported total at the evaluated sizes, so ninefold output density yields only a small increase in GMACs even though SCOPE is not the smallest model in parameter count. These tabulated counts characterize the specified arithmetic operations and are distinct from the relative-latency comparison below.

Figure~\ref{fig:compute_latency_trajectories} complements the arithmetic analysis with computation--latency trajectories at seven reconstruction factors from $2\times$ to $30\times$. SCOPE follows a comparatively narrow computation range while its latency increases with output density; LIIF-MS and LTE move toward both higher computation and higher latency. The near-vertical SCOPE trajectory therefore indicates limited arithmetic growth, not scale-independent execution time. EBCF-CDEM and EDSR retain lower latency at several plotted factors, so this comparison does not establish SCOPE as universally fastest. Together with the reconstruction results in Tables~\ref{tab:baseline_comparison} and~\ref{tab:scale_transfer_15x}, the trajectories support a joint assessment of reconstruction quality and scale-dependent execution cost. The $20\times$ and $30\times$ points extend the cost analysis only, not the reference-based accuracy evaluation.

\par\medskip
\noindent\begin{minipage}{\columnwidth}
\centering
\includegraphics[width=\linewidth]{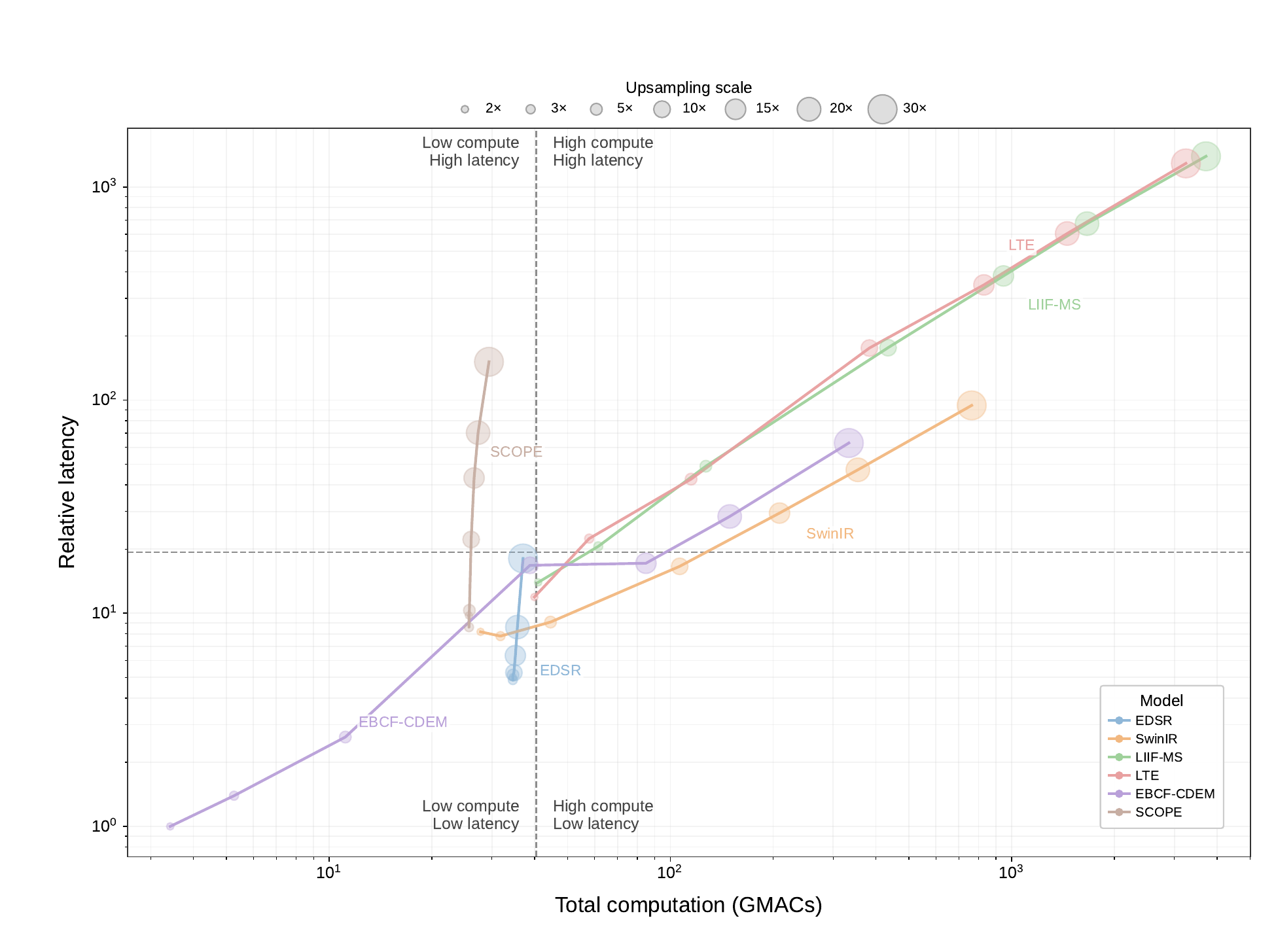}
\captionof{figure}{Computation--latency trajectories across reconstruction scales. Both axes are logarithmic; lower-left positions indicate lower cost. Marker sizes encode $2\times$, $3\times$, $5\times$, $10\times$, $15\times$, $20\times$, and $30\times$; lines connect increasing factors within each model. Dashed lines partition the cost space.}
\label{fig:compute_latency_trajectories}
\end{minipage}
\par\medskip

\subsection{Limitations and Extensions toward Latent Geomorphological Prior Learning}

SCOPE remains a paired grid-space reconstruction method, constrained by the information, noise, and uncertainty in the observed LR--HR products. The main quantitative comparisons use the validation subset employed for checkpoint selection, rather than a separate held-out main-study test set; the external marine experiment provides a distinct frozen-model evaluation. The present $15\times$ accuracy results cover land terrain because some regional 30~m marine products contain stronger local noise. Independently referenced 30~m bathymetric reconstruction is outside this evaluation. Small differences among training protocols and activations are reported descriptively, without a claim of statistical significance.

The geographic scope is bounded by the selected regions and sampling protocol. Polar and high-latitude terrain outside $65^\circ$N--$65^\circ$S is excluded because snow and ice cover, sparse observations, complex surface conditions, and vertical-reference differences introduce additional uncertainty. The external marine aggregate is patch-weighted and dominated by the Northern Australia and Great Barrier Reef groups, rather than equally weighted by source or representative of global bathymetry. Regional holdout does not establish independence of the underlying surveys incorporated into different bathymetric compilations. Similarly, the source-product extents and tile-index boundaries in Fig.~\ref{fig:study_area} indicate geographic context, not uninterrupted valid observations or complete tested coverage.

Grid spacing and evaluation metrics also delimit interpretation. The common 90~m external grid is a processing convention rather than a guarantee of effective 90~m bathymetric detail. Slope-related metrics use unit pixel spacing and assess grid-space structure rather than physical slope errors at 90~m or 30~m horizontal spacing. Fine-scale reconstruction remains an inference constrained by LR observations and learned terrain priors; it should not be interpreted as newly observed topographic measurements.

A more fundamental extension would be to move beyond grid-space sharpening toward latent geomorphological prior learning. LR and HR DEMs could be jointly embedded into a shared high-dimensional terrain manifold, where cross-scale consistency, basis-structured terrain variation, slope--aspect geometry, and terrain-type semantics are explicitly constrained. This direction should be distinguished from directly adopting codebook-based or diffusion-based image super-resolution: although learned latent priors have reduced ambiguity in restoration tasks~\cite{van2017neural,esser2021taming,zhou2022codeformer,gu2022vqfr,saharia2021image}, DEM reconstruction requires stronger physical and geomorphological control. Future work may therefore explore latent terrain-atlas learning, cross-scale representation alignment, uncertainty-aware constraints, and multi-source geospatial priors.

\section{Conclusion}
\label{sec:conclusion}

This study addresses continuous DEM reconstruction when paired training data are available at a lower reconstruction factor but labels at the desired target factor are unavailable for training. SCOPE learns a reusable coefficient field from the available pairs and reconstructs a denser elevation grid through local basis evaluation and geometry-guided ensemble fusion. The same design reduces the high-dimensional prediction repeated across output coordinates, addressing target-scale supervision constraints and computational growth together.

Under fixed $5\times$ supervision, SCOPE ranks first across six metrics in the main supervised-scale evaluation. In the land evaluation at the unseen $15\times$ factor, it reduces RMSE and MAE by approximately 12\% relative to bicubic, with RMSE within 0.08\% of target-scale fine-tuning. Ninefold output density increases the counted computation by approximately 2\%. Together, these results support effective reconstruction beyond the supervised resolution with low incremental arithmetic cost.

Frozen-model validation on 3,825 held-out external marine patches complements the scale-transfer evidence. Relative to the DEM-specific EBCF-CDEM baseline, SCOPE reduces RMSE by approximately 19\% under self-downsampling and 2\% with matched GEBCO\_2023 inputs, and yields lower RMSE than LIIF-MS in both settings. It leads all six metrics under self-downsampling and the learned models in RMSE, MAE, Slope, and PSNR under matched cross-product inputs. Together, the supervised-scale, unseen-scale, and external evaluations support reusable terrain functions as an effective approach to scale-flexible DEM reconstruction under limited target-resolution supervision.

\section*{Acknowledgements}
The authors gratefully acknowledge the German Aerospace Center (DLR) for providing the TanDEM-X 30~m EDEM, the GEBCO Compilation Group for the GEBCO\_2024 and GEBCO\_2023 grids, NOAA's National Centers for Environmental Information (NCEI) for the Coastal Relief Models, EMODnet Bathymetry for the EMODnet DTM, the Canadian Hydrographic Service for NONNA bathymetry, and Geoscience Australia and the AusSeabed contributors for the regional bathymetric products and MH370 data. This work was supported by the National Natural Science Foundation of China [grant number 42401446].

\section*{Declarations}
\hypersetup{bookmarksdepth=4}

\subsection*{Funding}
This work was supported by the National Natural Science Foundation of China [grant number 42401446].

\subsection*{Competing Interests}
The authors have no relevant financial or non-financial interests to disclose.

\subsection*{Author Contributions}
Zekai Shi: Conceptualization, Methodology, Software, Formal analysis, Investigation, Visualization, Writing---original draft. Meng Zhang: Supervision, Resources, Project administration, Writing---review and editing. Haokun Zhang: Validation, Writing---review and editing. Bo Zhang: Writing---review and editing. All authors read and approved the final manuscript.

\subsection*{Data Availability}
The datasets analyzed during the current study are publicly available from their original providers. Dataset names and access information are provided in the manuscript.

\bibliographystyle{elsarticle-harv}
\bibliography{references}

\end{document}